%% file: main.tex
\documentclass[a4paper, 10pt, conference]{ieeeconf}    
\usepackage{FG2026}

\FGfinalcopy

\input{preamble}

\IEEEoverridecommandlockouts                          
\def\FGPaperID{11} 

\title{\LARGE \bf
Zero-Shot Temporal Action Localization Through Textual Guidance
}

\author{\parbox{16cm}{\centering
    {\large Benedetta Liberatori$^1$, Alessandro Conti$^1$, Lorenzo Vaquero$^2$, Paolo Rota$^1$, Yiming Wang$^2$, Elisa Ricci$^1$}\\
    {\normalsize
    $^1$ University of Trento, Trento, Italy\\
    $^2$ Fondazione Bruno Kessler, Trento, Italy}}
}

\begin{document}

\ifFGfinal
\thispagestyle{empty}
\pagestyle{empty}
\else
\author{Anonymous FG2026 submission\\ Paper ID 11 \\}
\pagestyle{plain}
\fi
\maketitle

\input{sec/0_abstract}

\input{sec/1_introduction}

\input{sec/2_related}

\input{sec/3_method}

\input{sec/4_experiments}
\input{sec/conclusion}

\clearpage
\raggedbottom
{
    \small
    \bibliographystyle{ieee}
    \bibliography{main}
}

\renewcommand{\thesection}{\Alph{section}}
\input{supsec/supplementary}

\end{document}

%% file: preamble.tex
\usepackage[pagebackref,breaklinks]{hyperref}

\usepackage{tcolorbox}
\usepackage{verbatim}
\usepackage{amsmath}
\usepackage{colortbl} 
\usepackage{graphicx}
\usepackage{multirow}  
\usepackage{pifont}
\usepackage{tikz}
\usepackage{cancel}
\usepackage{rotating}
\usepackage[dvipsnames]{xcolor}
\usepackage{lineno}
\usepackage{booktabs}
\usepackage{xspace}
\usepackage{cleveref}
\usepackage{amssymb}
\usepackage{subcaption}

\newcommand{\ie}{\textit{i.e.}\xspace}
\newcommand{\eg}{\textit{e.g.}\xspace}

\newcommand{\inlineColorbox}[2]{\begingroup\setlength{\fboxsep}{1pt}\colorbox{#1}{\hspace*{2pt}\vphantom{Ay}#2\hspace*{2pt}}\endgroup}
\definecolor{ShadedGray}{RGB}{238,238,238}
\definecolor{ModelLightBlue}{RGB}{209, 233, 246}
\definecolor{DrawioBlue}{RGB}{218,232,252}
\definecolor{OurMethodColor}{RGB}{221, 241, 244}
\definecolor{DrawioOrange}{RGB}{255,230,204}
\definecolor{DrawioGreen1}{RGB}{204,255,204}
\definecolor{DrawioGreen}{RGB}{213,232,212}
\definecolor{DrawioPurple}{RGB}{225,213,231}
\definecolor{DrawioRed}{RGB}{248,206,204}
\definecolor{DrawioYellow}{RGB}{255, 242, 207}
\definecolor{DrawioBrightBlue}{RGB}{192,213,255}
\definecolor{DrawioBrightOrange}{RGB}{243,199,141}
\definecolor{DrawioBrightPurple}{RGB}{192,134,216}
\definecolor{MoreVividModelGreen}{RGB}{130, 200, 120}

\newcommand{\methodnameFull}{``\textbf{Te}xtual \textbf{Gu}idance for finer localization of actions in videos''\xspace}

\newcommand{\methodname}{\textsc{TeGu}\xspace}

\def\suppmat{\textit{Supp. Mat.}}

\newcommand{\refbase}[1]{\textcolor{cvprblue}{#1}} 
\definecolor{LineBlue}{RGB}{045, 114, 255}

\definecolor{cvprblue}{rgb}{0.21,0.49,0.74}

%% file: sec/0_abstract.tex
\begin{abstract}
Zero-shot temporal action localization (ZS-TAL) consists of classifying and localizing actions in untrimmed videos, where action classes are unseen at training time. Existing work uses Vision and Language Models (VLMs), taking advantage of their strong zero-shot transfer capabilities. Yet, these models face evident challenges with fine-grained action classification, making it difficult to directly use them to distinguish between the presence and absence of an action. 
Most current methods for ZS-TAL address these challenges by training models on large-scale video datasets, which require annotated data and often result in limited generalization performance. 
Recently, approaches discarding the use of labeled data have emerged as an alternative.
Following this direction, we propose a novel approach, \methodnameFull (\methodname), that compensates for the lack of supervision from training data by exploiting rich textual information derived from large language models and structured text extracted from captions. 
This additional linguistic context can improve fine-grained discrimination by providing richer cues about fine-grained action differences within videos. We validate the effectiveness of the proposed method by conducting experiments on the THUMOS14 and the ActivityNet-v1.3 datasets. Our results show that, by exploiting rich textual information for improved action localization, \methodname outperforms state-of-the-art ZS-TAL approaches that do not involve training
\footnote{Code available at \href{https://github.com/benedettaliberatori/tegu.git}{https://github.com/benedettaliberatori/tegu}}.
\end{abstract}

%% file: sec/1_introduction.tex
\section{Introduction}

Temporal Action Localization (TAL) in long, untrimmed videos remains a challenging and open problem in computer vision.
Traditionally, TAL has been approached through fully supervised methods, which rely on annotated training data for accurate action recognition and temporal localization~\cite{Zhang2022,Chao_2018_CVPR,Xu_2017_ICCV,sstad_buch_bmvc17,shou2017cdc,chao2018rethinking}. 
However, advancements in Vision and Language models (VLMs)~\cite{radford2021learning,li2023blip,NEURIPS2022_960a172b,xu-etal-2021-videoclip} have led to an increased interest in tackling TAL from a zero-shot perspective (ZS-TAL), where models aim to generalize to new actions without prior specific training on them. 
Recent works in this category \cite{Ju2021PromptingVM,stale,Yan_2023_ICCV,phan2024zeetad} incorporate zero-shot capabilities at inference time but still leverage in-domain knowledge through model fine-tuning on annotated data. 

\begin{figure}[t!]
\includegraphics[width=\linewidth]{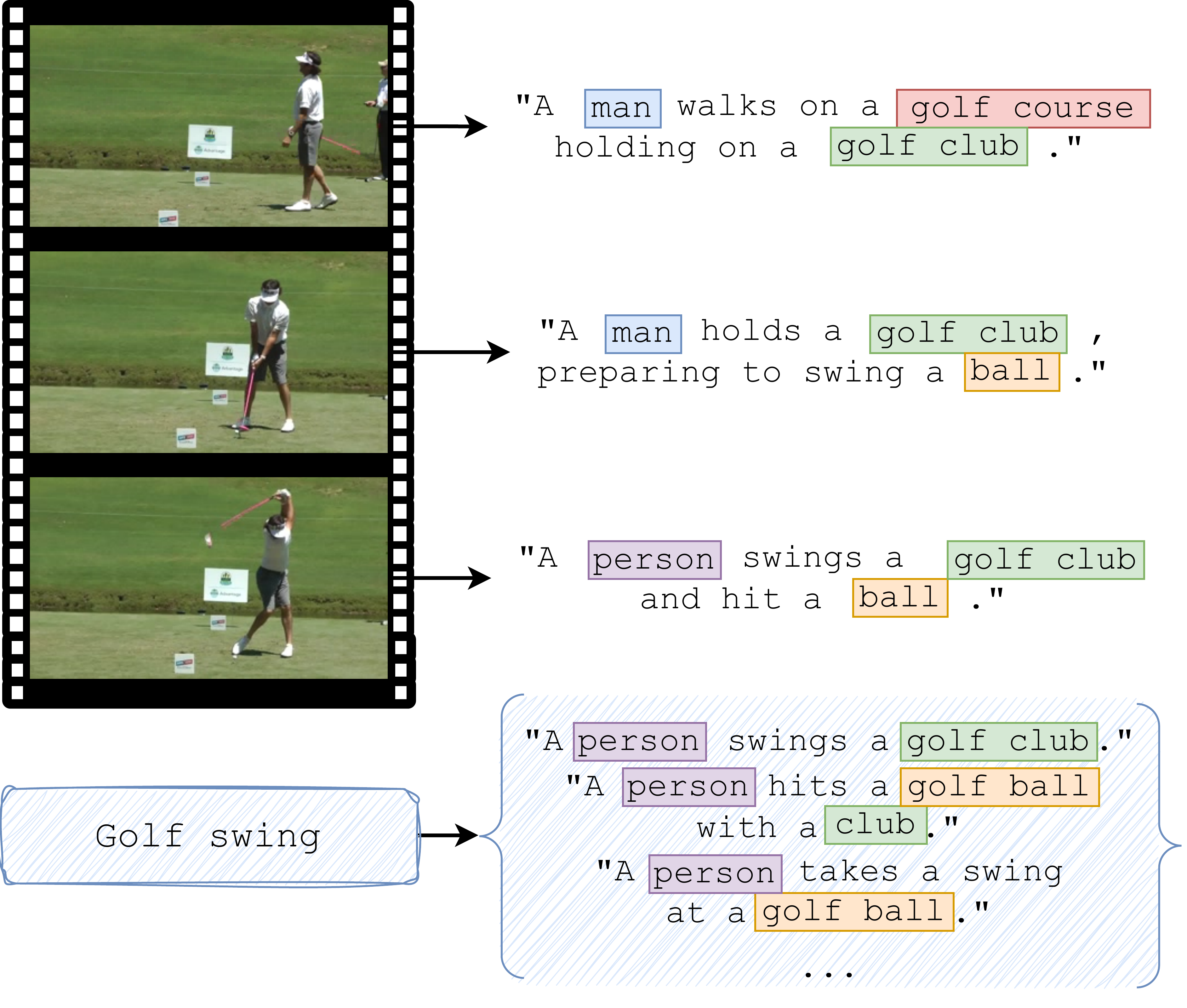}
\caption{
\textbf{Textual guidance for action localization.} We propose to use automatically extracted textual cues as an alternative to supervisory signal to adapt a pre-trained vision and language model to the task of temporal action localization. After generating class-level and video-level textual cues, we demonstrate the effectiveness of exploiting their combination to tune a model at test time.}

\label{fig:teaser}
\end{figure}
Increased supervision usually leads to better results; however, the cost of manual annotation can be prohibitive. 
Additionally, certain applications may impose restrictions on data storage for training purposes.
A more recent line of work addresses these challenges by proposing a ZS-TAL setting in which training data is not required.
Leveraging only a single video and introducing a carefully designed frame selection strategy along with the corresponding objective function, T3AL~\cite{Liberatori2024} improves the model's performance directly during inference.
While this approach incorporates some form of textual supervision, it is confined to a supporting role rather than being central to the model’s understanding.

In this work, we explore the potential of language-based cues to address ZS-TAL at inference time. Our motivation stems from the extensive usage and promising results of textual information in complex video understanding tasks, such as those tackled in~\cite{tao2024navero,Momeni2023,li2024unimotion}, which improved model comprehension of videos featuring complex scenarios.

A common strategy in video understanding is to supplement frame information with captions describing the visual content. 
While effective for understanding short, focused videos, our study suggests that captions alone introduce substantial noise in the temporal localization task (see Fig.~\ref{fig:teaser}), where \textit{background} frames (\ie, not depicting the action) are often described similarly to those highlighting \textit{foreground} frames, making captions an unreliable signal. 
To address this limitation, we propose \methodnameFull (\methodname), a novel approach leveraging textual guidance in the form of video-level \textit{scene triplets} and action \textit{class descriptors}. 
Specifically, for each frame in the video, we generate a caption and utilize a scene-graph text parser to process it, obtaining scene triplets in the form of $<$\texttt{subject}, \texttt{verb}, \texttt{object}$>$. 
Our intuition is that scene triplets offer a more structured and manipulable element to identify key moments within videos. 
They not only allow multiple and distinct semantic cues to be drawn from each frame but also emphasize verbs (\ie, critical indicators of actions), while preserving contextual information on the actors and objects involved. 
Moreover, we extract information directly from class names by generating descriptions and listing action-related objects with a Large Language Model (LLM). 

Inspired by~\cite{Liberatori2024}, we propose to integrate the textual information derived from scene triplets and LLM-enriched action descriptions into a novel test-time adaptation framework.
Among the scene triplets, we first identify those that are semantically closer to the action class (\ie, \textit{affine triplets}) and those that are farther away (\ie, \textit{distractor triplets}) to compute a similarity-based score for localizing the action intervals. This score is then refined by accounting for the similarity between frame-based visual representations and textual action descriptions.  We then adopt a similar adaptation strategy at inference time, considering all the frames of a single video as in~\cite{Liberatori2024}.
Differently, our process benefits from textual information and is supervised with a max-margin loss which imposes large separation between scores corresponding to action and background frames.
Our experiments on the challenging THUMOS14~\cite{IDREES20171} and ActivityNet-v1.3~\cite{7298698} datasets demonstrate that \methodname achieves state-of-the-art performance when compared to previous methods which do not require training. Our results confirm the benefit of integrating linguistic cues for enhancing model's performance in ZS-TAL.

This work provides the following contributions:

\begin{itemize}
  
\item We introduce \methodname, a novel framework that effectively leverages textual information from image captions and Large Language Models to improve ZS-TAL on a single video to adapt a pre-trained image-text model to ZS-TAL.  

\item We propose an innovative strategy to integrate linguistic cues, derived from scene triplets and semantically-enriched action descriptions, into an adaptation framework, guided by a max-margin loss for supervision.

\item Our approach is complemented with an in-depth analysis that highlights the challenges and provides insights of incorporating linguistic information into ZS-TAL. 
\end{itemize}

%% file: sec/2_related.tex
\section{Related work}

\subsection{Temporal Action Localization.}
Temporal Action Localization involves detecting and recognizing actions within videos.
Traditional approaches are typically divided into two categories: two-stage and one-stage methods.
Two-stage methods~\cite{Qing2021,Li2024, Xu_2017_ICCV, Chao_2018_CVPR, Ju2021PromptingVM} first generate class-agnostic action proposals and subsequently classify each proposal individually.
In contrast, one-stage methods~\cite{Zhang2022, lin_acm, stale, Yan_2023_ICCV,Liu2024} integrate localization and classification simultaneously into a unified process.
Fully supervised TAL approaches~\cite{Zhang2022,Chao_2018_CVPR,Xu_2017_ICCV,sstad_buch_bmvc17} require precise temporal annotation of each action instance, which is labor-intensive and impractical. 
In contrast, weakly supervised approaches~\cite{wang2017untrimmednets,he2022asm,yang2021uncertainty,Ju2023} require only video-level category labels for training.
Both commonly operate under a closed-set assumption, meaning the action categories in the training set are the same as those in the test set.
For this reason, zero-shot (ZS-TAL)~\cite{Ju2021PromptingVM,stale,Yan_2023_ICCV,phan2024zeetad} and open-vocabulary (OV-TAL)~\cite{10517407,Hyun_2025_WACV,Gupta_2024_BMVC} settings have been proposed, which allow models to recognize unseen action categories during inference.
Specifically, ZS-TAL ensures separation between seen and unseen action categories during training and evaluation, while OV-TAL evaluates models on both seen and unseen categories.
We tackle the more restrictive scenario, previously introduced in~\cite{Liberatori2024}, where no training data is available, making the ZS-TAL setting closer to ours as it evaluates models on entirely new categories.

\subsection{Zero-Shot Temporal Action Localization.}
EffPrompt~\cite{Ju2021PromptingVM} introduces the ZS-TAL setting and employs a two-stage architecture that uses an action proposal generator~\cite{Lin_2021_CVPR}. This detector identifies potential action segments, which are then classified using text-image embeddings from CLIP~\cite{radford2021learning}.
STALE~\cite{stale} presents a proposal-free, CLIP-based architecture for ZS-TAL that uses two parallel streams to learn relevant localization masks and classify these regions according to their alignment with action classes.
Similarly, UnLoc~\cite{Yan_2023_ICCV} extracts joint image-text features using CLIP, processes them with a fusion module, and applies a feature pyramid network to predict per-frame relevance scores and define action boundaries.
ZEETAD~\cite{phan2024zeetad} proposes a dual-localization mechanism fine-tuned on action recognition data to perform zero-shot temporal action detection.
While these methods improve ZS-TAL, they all rely on annotated data for training, which may be unavailable or challenging to store in real-world scenarios.
Furthermore, T3AL~\cite{Liberatori2024} points out that existing ZS-TAL methods face challenges in cross-dataset scenarios. Models trained on one dataset perform poorly when tested on another one due to biases introduced during training, such as variability in action length or frequency.
To address this, T3AL~\cite{Liberatori2024} proposes a \textit{fully unsupervised ZS-TAL setup where no training dataset is available},
leveraging test-time adaptation and frame captioning.
Despite these innovations, existing approaches still underutilize textual information, treating it as a complementary rather than a primary feature. This limitation suggests an opportunity to further explore textual information in unsupervised settings for more effective ZS-TAL. \methodname aims to investigate this largely unexplored research direction.

\subsection{Test-Time Adaptation.}
Model optimization is widely used to adapt pre-trained models to changes in data distribution. Test-time adaptation (TTA) methods require only the pre-trained model and unlabeled test data for optimization during inference. 
Among these, TENT~\cite{wang2021tent} adapts the model by minimizing the entropy of its predictions on a batch of test samples.
While TENT requires more than one sample to get a non-trivial solution, MEMO~\cite{NEURIPS2022_fc28053a} augments a single test sample and minimizes the entropy of the marginal distribution across augmentations.
More recently, other work has repurposed TTA techniques to improve the robustness and generalization of VLMs~\cite{NEURIPS2023_cdd06402,shu2022tpt,samadh2023align}.
Although most of this research has focused on image-level tasks, some recent work has extended these approaches to video action recognition, adapting spatio-temporal video models to improve robustness to video corruption ~\cite{lin2023video,yi2023temporal} and generalization~\cite{xiong2024modality}. 
ViTTA first generates augmented views of the input videos by temporally resampling frames, and then it minimizes the difference between training statistics and online estimates of test statistics.
TeCo~\cite{yi2023temporal} minimizes the prediction entropy on sparse video streams while adding a temporal coherence regularization term on dense video streams.
MC-TTA~\cite{xiong2024modality} initializes a teacher and a student model from the pre-trained model, with memory banks for generating pseudo-prototypes and target-prototypes, respectively. Then, it mitigates domain shift
by reducing the difference between the pseudo-source and target distributions.
These TTA methods focus on adapting a pre-trained model to handle distribution shifts while staying within the same fundamental task, \ie image or video classification, making them non-trivially applicable at inference time to a different video task, such as temporal action localization.

%% file: sec/3_method.tex
\section{Method}\label{sec:methodParent}
This section provides a formal introduction to the Zero-Shot Temporal Action Localization (ZS-TAL) task (\cref{sec:pf}) and details our proposed approach (\cref{sec:method}), breaking down each of its key components.

\subsection{Problem Formulation}
\label{sec:pf}

\begin{figure*}[t!]
\makebox[\linewidth][c]{\includegraphics[width=1.0\textwidth]{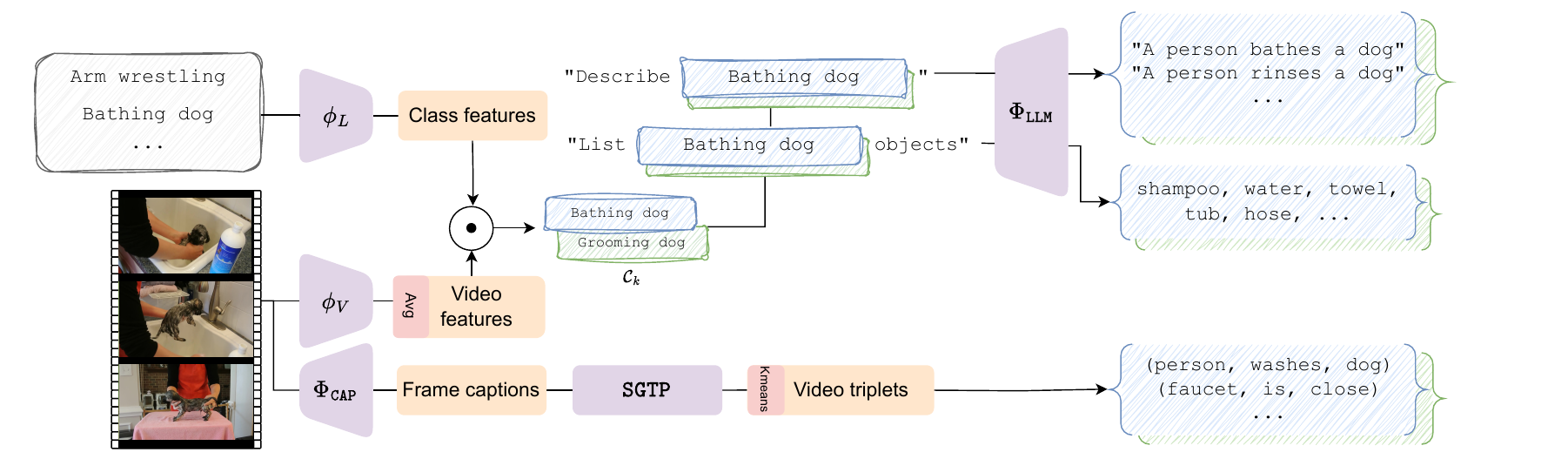}}
\caption{\textbf{\methodname steps 1-2.} Given an input video, \methodname first \inlineColorbox{DrawioPurple}{encodes} the class names and its frames into \inlineColorbox{DrawioOrange}{class features} and its \inlineColorbox{DrawioRed}{average} \inlineColorbox{DrawioOrange}{video features} to get the \inlineColorbox{DrawioBlue}{class predictions} at video level. Then, it \inlineColorbox{DrawioPurple}{generates} the predicted class' \inlineColorbox{DrawioBlue}{descriptions} and \inlineColorbox{DrawioBlue}{objects}. Concurrently, it \inlineColorbox{DrawioPurple}{generates} \inlineColorbox{DrawioOrange}{frame captions} and \inlineColorbox{DrawioPurple}{parses} scene graphs, which are \inlineColorbox{DrawioRed}{clustered} to remove redundancy and get \inlineColorbox{DrawioOrange}{video triplets}. 
}
\label{fig:method}
\end{figure*}

\begin{figure}[t!]
\makebox[\linewidth][c]{\includegraphics[width=0.5\textwidth]{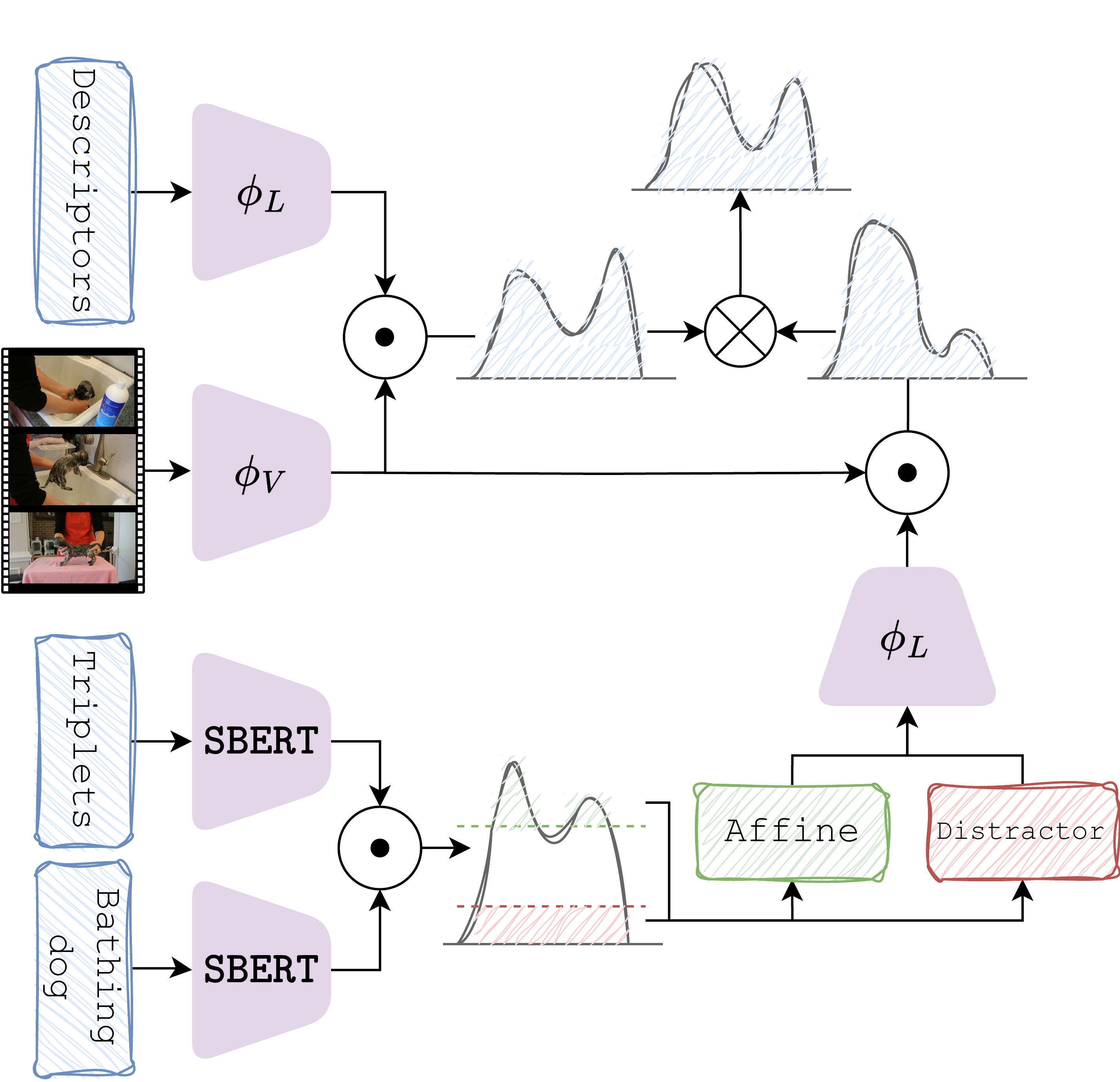}}
\caption{\textbf{\methodname step 3.} Using the \inlineColorbox{DrawioBlue}{text guidance} from previous steps, \methodname \inlineColorbox{DrawioPurple}{encodes} and estimates a first frame-wise score for localization, which is then refined using \inlineColorbox{DrawioBlue}{video triplets} split into \inlineColorbox{DrawioGreen}{affine triplets} and \inlineColorbox{DrawioRed}{distractor triplets}.
}
\label{fig:method2}
\end{figure}
Given a test video $\mathcal{V}=\{x_i\}_{i=1}^N$ comprising $N$ frames $x_i \in \mathcal{X}$ and a set of $Z$ action categories $\mathcal{C}=\{c_z\}_{z=1}^Z$, TAL aims to temporally localize and classify a set of $M$ actions $\{(t_j,y_j)\}_{j=1}^M$ in the video $\mathcal{V}$, where $t_j = (a_j, b_j) \in \mathbb{R}^2$ denotes the start and end timestamps of each action, and $y_j \in \mathcal{C}$ specifies its category. 
In the zero-shot setting, the categories $\mathcal{C}$ are assumed to be unseen during training, using a labeled training dataset with alternative categories $\mathcal{C}^{train}$ such that $\mathcal{C}^{train} \cap \mathcal{C} = \varnothing$.
Unlike previous works~\cite{Ju2021PromptingVM,phan2024zeetad,stale,Yan_2023_ICCV} that train a model using annotated in-domain data, we address a more practical and less explored ZS-TAL setting where in-domain training data $\mathcal{C}^{train}$ are unavailable~\cite{Liberatori2024}. Only one single unlabeled video is used at inference time.

\subsection{Textual Guidance for ZS-TAL}
\label{sec:method}
Our approach, \methodname, leverages textual guidance from captioning and language models to adapt a pre-trained VLM for ZS-TAL.
\methodname operates in four main steps:
In \textbf{Step~1} it uses a VLM to identify the main actions occurring in the video.
In \textbf{Step~2}, it generates frame-level captions and carefully parses them into scene triplets to extract key details. Moreover, it generates textual descriptors of the main actions identified with an LLM (see Fig.~\ref{fig:method}).
Next, in \textbf{Step~3} it uses the extracted textual information to differentiate between frames that are likely to depict the main actions and those that are not relevant (see Fig.~\ref{fig:method2}).
Finally, in \textbf{Step~4}, the selected frames are utilized to adapt the VLM in an unsupervised manner, enhancing its discriminative ability. 
\newline
 
\noindent \textbf{Step 1: Video-level action classification.} 
Given a VLM comprising a vision encoder $\phi_V$ and a language encoder $\phi_L$, we perform video-level classification to identify the top-$k$ main actions $\mathcal{C}_k \subset \mathcal{C}$ depicted in $\mathcal{V}$. 
First, we compute the cosine similarity with the classes, averaged over the $N$ video frames:
\begin{equation}
\sigma(c | \mathcal{V}) = \cos \left( \frac{1}{N} \sum_{i=1}^N \phi_V(x_i), \phi_L(c)\right) \quad \forall c\in\mathcal{C},
\end{equation}
where $\cos(\cdot, \cdot)$ denotes the cosine similarity.
These classification scores $\sigma$ are used to rank all candidate actions, and then the top-$k$ actions with the highest scores are selected.
Given the predicted set of actions $\mathcal{C}_k$, we aim to identify the intervals within the video $\mathcal{V}$ that contain these actions.
\newline

\noindent \textbf{Step 2: Textual guidance extraction.}
Although the class names are sufficient to identify the main actions happening in $\mathcal{V}$, they are far too coarse to distinguish between frames where each $c_k\in\mathcal{C}_k$ is taking place and where it is absent.
They also lack the necessary details to effectively differentiate between semantically similar activities that may precede or succeed the staging of an action, \eg, ``\textit{running}" before a ``\textit{high jump}", as current VLMs tend to exhibit a strong bias toward objects, often neglecting verbs~\cite{Momeni2023,NEURIPS2023_42049206}. To address this limitation, we enrich the textual information with action-specific cues from both class-level and frame-level descriptions.

Regarding the class, we use an LLM, denoted as $\Phi_{\texttt{LLM}}$, to enrich the semantic details of each $c_k\in\mathcal{C}_k$ by generating detailed action descriptions and by listing the main objects involved in its execution.
Specifically, given a prompt $P_a$ for the generation of detailed action descriptions and a prompt $P_o$ for the generation of main objects involved in the action, we can create for each $c_k\in\mathcal{C}_k$ two sets: $\mathcal{D}^a_{c_k} = \Phi_{\texttt{LLM}}(P_a, c_k)$ and $\mathcal{D}^o_{c_k} = \Phi_{\texttt{LLM}}(P_o, c_k)$, both containing semantically-rich texts related to action $c_k$.
\methodname will use these action-specific descriptions as its primary textual guide for ZS-TAL.
For more details on the specific prompts used in our implementation, as well as examples of the contents of $\mathcal{D}^a_{c_k}$ and $\mathcal{D}^o_{c_k}$, refer to the \suppmat

Regarding the frames,
we first leverage a captioning model, denoted as $\Phi_{\texttt{CAP}}$, to generate a set of captions 

\begin{equation}
\mathcal{G}~=~\{ g_i~=~\Phi_{\texttt{CAP}}(x_i) | x_i \in \mathcal{V} \},
\end{equation}
 originating from each frame in the video sequence $\mathcal{V}$.
Next, we apply a scene-graph text parser, denoted as $\texttt{SGTP}$,
to extract scene triplets. 
Given a caption $g_i\in \mathcal{G}$, we parse a set of
textual triplets, in the form $\texttt{<s, v, o>}$, where \texttt{s}, \texttt{v}, and \texttt{o} represent the subject, verb, and object, respectively.
We extract these triplets for each frame, as 
\begin{equation}
\mathcal{T_\mathcal{V}}~=~\{ \texttt{SGTP}(g_i) | g_i \in \mathcal{G} \}.    
\end{equation}

Since consecutive frames often display similar visual content, producing redundant information, we filter out repetitive triplets in $\mathcal{T}_\mathcal{V}$.
Specifically, we evaluate the textual embedding for each triplet with SentenceBERT~\cite{reimers2019sentence}, and perform k-means clustering~\cite{1056489} to retain only semantically different triplets.
The resulting set of $S$ scene triplets, $\mathcal{T}^S_{\mathcal{V}}$, provides a concise summary of the primary content in $\mathcal{V}$, capturing the essential events of the video in a compact and structured format.
\newline

\noindent \textbf{Step 3: Score calculation for action localization.} 
This step aims to identify the temporal intervals of each action $c_k\in\mathcal{C}_k$ by computing a frame-level score $s_i \in [0, 1]$, where the higher the value of $s_i$, the more likely it is that the frame $x_i$ depicts an action of category $c_k$.
While we will use only $c_k$ in the notation for simplicity, the following steps apply equally to all identified actions $c_k \in \mathcal{C}_k$.
To compute the score $s_i$, we leverage the textual cues extracted in the previous step and the image and text encoders from the VLM.

Specifically, using the LLM-enriched action descriptions $\mathcal{D}_{c_k} = \left\{ \mathcal{D}_{c_k^a},~\mathcal{D}_{c_k^o} \right\}$ associated to action category ${c_k}$, we estimate for each frame $x_i$ an initial localization score $s_i$ as: 
\begin{equation}
    s_i = \sum_{d\in \mathcal{D}_{c_k}}\pi(x_i, d),
\label{eq:score}
\end{equation}
where $\pi\left(\cdot,\cdot\right)$ is the probability indicating the alignment between frames and descriptions, as computed by the VLM.

The score $s_i$ derived from the LLM-enriched action descriptions may not be sufficient to distinguish the action's foreground (\ie, the frames where the action is taking place) from the background, as it lacks semantic grounding to the specific video. 
We further propose to employ scene triplets $\mathcal{T}_\mathcal{V}^S$ that are extracted from the video to better calibrate the scores $s_i$ computed for each frame $x_i$ to the given video $\mathcal{V}$. 

Specifically, we categorize the scene triplets $\mathcal{T}^S_\mathcal{V}$ into two groups: \textit{affine} triplets $\mathcal{T}^a_\mathcal{V}$, which exhibit a high semantic relevance with the action class $c_k$, and \textit{distractor} triplets $\mathcal{T}^d_\mathcal{V}$, which may be just weakly relevant to the context of target class. 
For instance, given the class indicating the action ``\texttt{bathing dog}", an example of an \textit{affine} triplet is \texttt{<person, wash, dog>}, while an example of a \textit{distractor} triplet is \texttt{<faucet, is, close>}.

This grouping is achieved by computing the cosine similarity between the SentenceBERT~\cite{reimers2019sentence} embeddings of each triplet in $\mathcal{T}^S_\mathcal{V}$ and the action class name. 
Thus, we form the sets $\mathcal{T}^a_\mathcal{V}$ and $\mathcal{T}^d_\mathcal{V}$, by selecting the top-$k$ and bottom-$k$ scene triplets according to cosine similarities.
With $\mathcal{T}^a_\mathcal{V}$ and $\mathcal{T}^d_\mathcal{V}$, we then refine the score $\bar{s}_i$ for each frame $x_i$ as:
\begin{equation}
    \bar{s}_i = s_i + \alpha \left( \sum_{a\in \mathcal{T}^a_\mathcal{V}}\pi\left(x_i, a\right) - \sum_{d\in \mathcal{T}^d_\mathcal{V}}\pi\left(x_i, d\right) \right),
\end{equation}
where \( \alpha \) is a weighting factor that balances the influence of the contribution of the scene triplets and the descriptors to the final scores $\bar{s}_i$.
Overall, as shown in our experiments (Sec.~\ref{sec:sota}), the combined use of action-specific descriptions ($\mathcal{D}_{c_k}$) and affine/distractor scene triplets ($\mathcal{T}^a_\mathcal{V}$ and $\mathcal{T}^d_\mathcal{V}$) lead to improved action localization performance. 
\newline

\noindent \textbf{Step 4: Model adaptation.} 
In the last step of our method, given the video $\mathcal{V}$, we propose to fine-tune a subset of parameters of the model to improve TAL. For the purpose, we leverage frames with high-confidence predictions as pseudo-labeled samples. 
Specifically, we construct two subsets from $\mathcal{V}$: the positive set $\mathcal{P}$, which includes frames with scores exceeding the threshold established by the top-$p$ percentile of scores $\bar{s}$, and the negative set $\mathcal{N}$, which consists of frames with scores below the threshold defined by the bottom-$p$ percentile of scores $\bar{s}$. 

The adaptation process aims to maximize the score separability between $\mathcal{P}$ and $\mathcal{N}$ such that the minimum score of the frames within $\mathcal{P}$ exceeds the maximum score within $\mathcal{N}$, formally expressed as:

\begin{equation}
    \min_{x_i\in \mathcal{P}}\bar{s}_i > \max_{x_i\in \mathcal{N}}\bar{s}_i.
\label{eq:loss}
\end{equation}

To enforce this condition, we apply a max-margin ranking loss that constrains the lowest positive score to be higher than the highest negative score by a margin $\gamma$: 
\begin{equation}
   \mathcal{L}_m = \max\left( 0, \gamma - \min_{x_i\in \mathcal{P}}\bar{s}_i + \max_{x_i\in \mathcal{N}}\bar{s}_i\right). 
\end{equation}

Moreover, we add a regularization term for temporal smoothness defined as:
\begin{equation}
\mathcal{L}_{tmp} = \frac{1}{N}\sum_{i=1}^N{\left \lVert s(x_i) - s(x_{i-1}) \right \rVert_2}.
\end{equation}
This term promotes temporal smoothness between the localization scores of temporally adjacent frames by minimizing the difference in the scores for adjacent video frames.
The overall objective function for model adaptation is:
\begin{equation}
\mathcal{L}=\mathcal{L}_m+\lambda\mathcal{L}_{tmp}, \label{eq:obj}
\end{equation}
where $\lambda$ is a regularization parameter.

To adapt the model, we minimize the objective function in Eq.~\ref{eq:obj} for $T$ steps (see implementation details in Sec.~\ref{sec:experiments}).
Then, we re-compute the scores with Eq.~\ref{eq:score} with the optimized encoders $\phi_L^\ast,$ and $\phi_V^\ast$. We convert the scores into action proposals using a threshold set to the average score across all frames of each video, as in~\cite{Liberatori2024}. 
To predict intervals from multiple action categories, we repeat the process for all the identified main actions $c_k\in\mathcal{C}_k$. 
In practice, if the top category in the video has a prediction confidence that exceeds a certain threshold, we focus solely on that top category. 
Finally, we use non-maximum suppression to refine the predicted intervals. 
After the final prediction on the input video $\mathcal{V}$, the model is re-initialized to its original pre-trained state for processing a new video.

%% file: sec/4_experiments.tex
\section{Experiments}
\label{sec:experiments}
We validate our proposed method, \methodname, on two widely used benchmarks for ZS-TAL and compare its performance against state-of-the-art methods. 
Additionally, we conduct an ablation study to justify our key design choices and evaluate the limitations associated with captions.
We first describe the experimental setup in terms of the datasets, settings, and metrics used, then present and discuss the results in Sec.~\ref{sec:sota}, the ablation study in Sec.~\ref{sec:ablation}, and the captions' evaluation in Sec.~\ref{subsec:captions}.

\noindent\textbf{Datasets.} We validate our proposed method on two temporally annotated untrimmed video datasets, THUMOS14~\cite{IDREES20171} and ActivityNet-v1.3~\cite{7298698}. THUMOS14 dataset contains 413 videos with 20 sports-related action categories, while ActivityNet-v1.3 is a large-scale dataset containing 19,994 videos with 200 action categories.

\noindent\textbf{Metrics and settings.} We measure the performance using the mean Average Precision (mAP) at different temporal IoU thresholds: [$0.3$:$0.1$:$0.7$] for THUMOS14 and [$0.5$:$0.05$:$0.95$] for ActivityNet-v1.3 datasets. 
Following prior work, we report the results across two different settings, where $75\%/50\%$ of the classes appear during training and the remaining $25\%/50\%$ are only seen at inference. Results are averaged on 10 random splits of classes. 

\noindent\textbf{Competitors.}
We compare \methodname against a mixture of training-based and training-free/test-time approaches.
The training-based competitors include EffPrompt~\cite{Ju2021PromptingVM}, STALE~\cite{stale}, UnLoc~\cite{Yan_2023_ICCV}, and ZEETAD~\cite{phan2024zeetad}, all of which utilize in-domain data for training relevant to the TAL task. 
Additionally, we include the two-stage baseline proposed in~\cite{Ju2021PromptingVM} that equips CLIP with a detector trained for the task~\cite{Lin_2021_CVPR,yang2020revisiting}.
These methods operate under different conditions and are included here primarily as a \textit{training-based reference} rather than for direct comparison.
Lastly, we identify T3AL~\cite{Liberatori2024} as the method closest to our setting, and consider it our main competitor as it shares a similar data policy, \ie, test-time adaptation without supervised signal from an image-text backbone.
We evaluate all the competitors on both datasets except for UnLoc, which was only trained and evaluated on ActivityNet-v1.3.

\noindent\textbf{Implementation Details.} 
We employ SigLIP-SoViT-400M/14~\cite{zhai2023sigmoid} as our VLM.
For instance-level text guidance, we use KOSMOS-2~\cite{peng2023kosmos} as the captioning model $\Phi_{\texttt{CAP}}$ and the scene-graph text parser $\texttt{SGTP}$ from~\cite{li-etal-2023-factual}.
For class-level text guidance, we use \textit{gemini-1.5-flash}~\cite{team2023gemini} as $\Phi_{\texttt{LLM}}$.
We decode frames at the full frame rate for THUMOS14 and at 1 FPS for ActivityNet-v1.3, resizing them to a $224\times224$ spatial resolution. We augment action class names with the template \text{\texttt{"A video of action \{\}"}}. We use AdamW optimizer with a learning rate of $1e^{-4}$ and set $\lambda=1e^{-2}$, adapting the last linear layer of $\phi_T$ and the last MLP of $\phi_V$ for $T=10$  steps on THUMOS14 and $T=3$ on ActivityNet-v1.3. 
Refer to \suppmat~for additional implementation details.

\subsection{Comparison with the state-of-the-art}
\label{sec:sota}

We compare \methodname with state-of-the-art ZS-TAL methods, encompassing both training- and non-training-based approaches. 
The results are presented in Tab.~\ref{thumos} and Tab.~\ref{anet}, which report performance metrics on THUMOS14 and ActivityNet-v1.3 for both the $50\%-50\%$ and $75\%-25\%$ settings.
We indicate with \textit{Detector + CLIP} the two-stage baseline consisting of a pre-trained detector and CLIP as the classifier defined in \cite{Ju2021PromptingVM}.  
\methodname outperforms the state-of-the-art results when compared to methods that do not use training data, demonstrating an improvement of $+1.8/0.1\%$ and $+3.6/4.2\%$ mAP on the two settings for THUMOS14 and  ActivityNet-v1.3, respectively. 

\input{tables/thumos}

\input{tables/anet}

\subsection{Ablation Study}\label{sec:ablation}
In this section, we perform ablations of our method to validate our main design choices: the textual guidance extraction, the model adaptation at test-time, the learning objective employed, and the number of centroids used for clustering scene triplets.  
We further conduct a dedicated analysis to evaluate the merits and downsides of the textual information, provided by the rich frame-level captions, on addressing ZS-TAL.
We perform the experiments on the THUMOS14 dataset and report numbers for both settings.

\noindent \textbf{Text guidance.}
In Tab.~\ref{table:text}, we evaluate the effectiveness of the action cues, as introduced in Sec.~\ref{sec:method}. We indicate with $\mathcal{D}$ and $\mathcal{T}$ the use of descriptors and scene triplets, respectively. The experiment shows that removing both components results in performance degradation, with mAP variation of $+2.8/2.8\%$ compared to the final method. 
The individual use of both components, however, enhances the adaptation of the model, leading to improved performance.

\noindent \textbf{Model adaptation.}
In Tab.~\ref{table:loss} we analyze the impact of adapting the model at test-time and the choice of the learning objective. 
We compare our objective function with the one proposed in~\cite{Liberatori2024}, denoted in the table as \textsc{BYOL}. The work adapted the BYOL loss~\cite{NEURIPS2020_f3ada80d} to push the scores of positive samples closer to $1$ and the one of negative close to $0$. 
In contrast, we denote the loss defined in Eq.~\ref{eq:loss} with $\mathcal{L}_m$.  
The results indicate that both loss functions improve performance relative to non-adapted models, but our $\mathcal{L}_m$ loss is more effective in increasing the margin between positive and negative samples. This demonstrates its superior ability to enhance score refinement compared to the \textsc{BYOL} one.

\noindent \textbf{Number of scene triplets.} 
We show in Fig.~\ref{fig:kmeans} the impact of varying the cardinality $S$ of the set of video-specific triplets. For both settings,  we can observe that changing the value of $S$ in small ranges does not result in significant variations in performance. However, as $S$ increases, performance declines, indicating that a bigger set of triplets corresponds to less effective score refinement.

\input{figures/clustering}
 
\input{tables/ablation0}
\input{tables/ablation2}

\label{sec:exp:text}
\begin{figure}[t!]
    \centering
    \includegraphics[width=\linewidth]{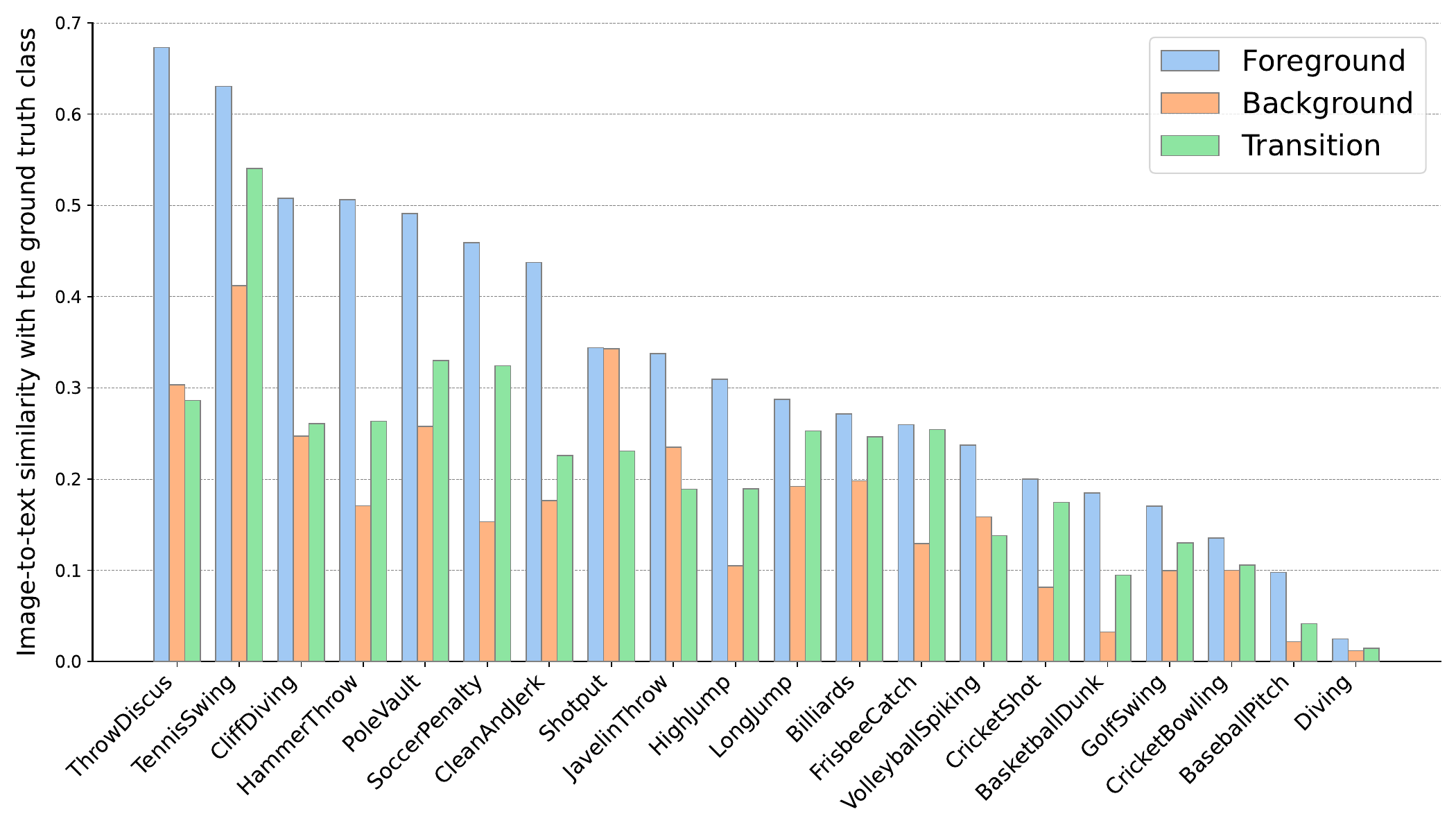}
    \caption{\textbf{Image-to-text similarities.} Cosine similarities between frames and ground truth video classes, computed on VLM embeddings. 
    Numbers are calculated for \inlineColorbox{DrawioBlue}{foreground}, \inlineColorbox{DrawioOrange}{background}, and \inlineColorbox{DrawioGreen}{transition} frames, and averaged across each class.
    \vspace{-10pt}}
    \label{fig:sim_vlm}
\end{figure}

\begin{figure}[t!]
    \centering
    \includegraphics[width=\linewidth]{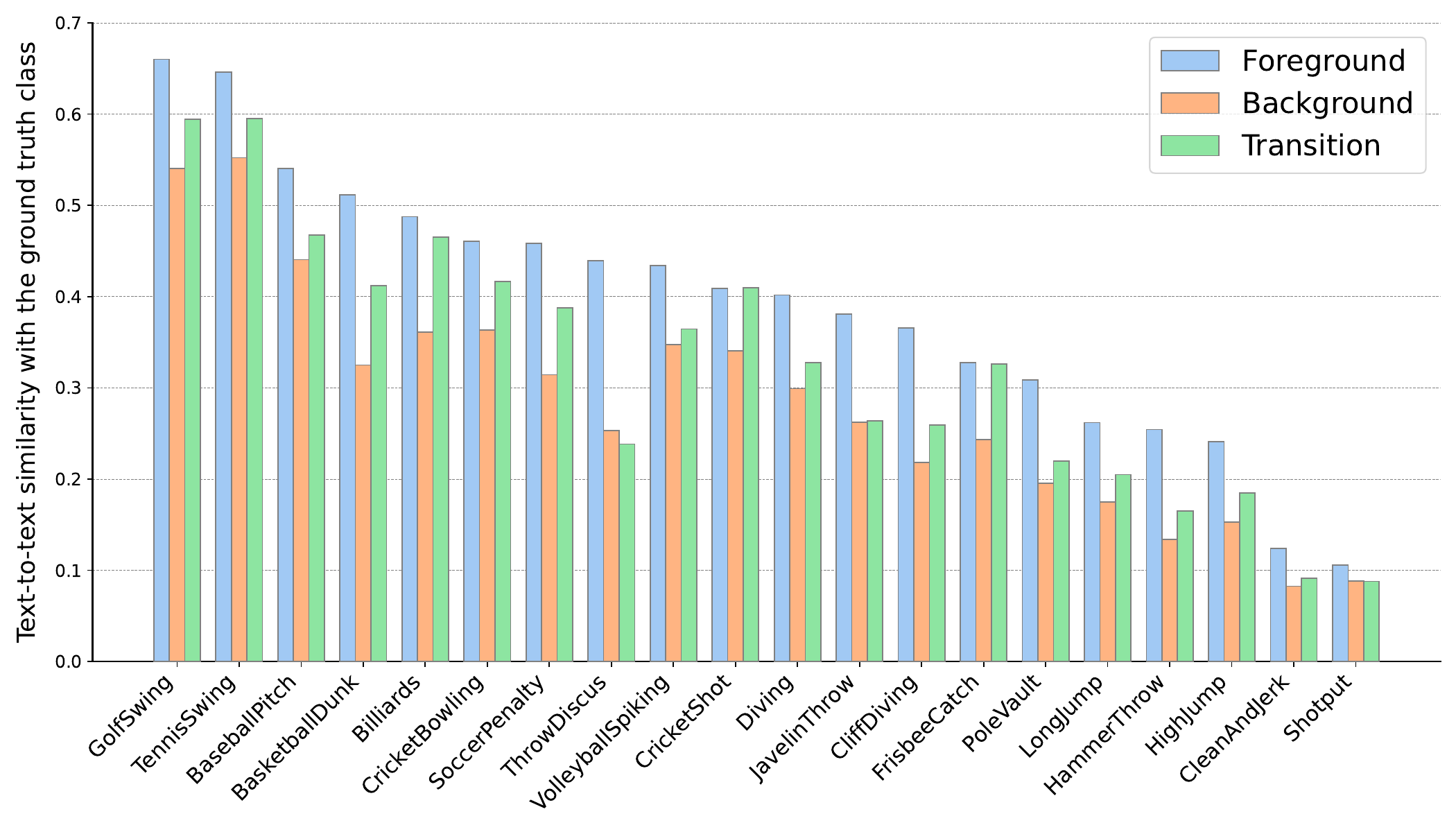} 
    \caption{\textbf{Text-to-text similarities.} Cosine similarities between frame-level captions and ground truth video classes, computed on SentenceBERT embeddings. 
    Numbers are calculated for \inlineColorbox{DrawioBlue}{foreground}, \inlineColorbox{DrawioOrange}{background}, and \inlineColorbox{DrawioGreen}{transition} frames, and averaged across each class.
    \vspace{-10pt}}
    \label{fig:sim_sbert}
\end{figure}

\subsection{On the caveat of captions for ZS-TAL}\label{subsec:captions}
We further investigate how captions, supposedly semantically rich, may impact ZS-TAL and, more importantly, how to better utilize captions for this task. 
We mainly focus on how their information can impact delineating the challenging \textit{transitional} intervals of actions, which have a crucial impact on ZS-TAL performance.
The transitions often capture the before-and-after phases of actions, thus being semantically ambiguous to both the action and the background context.

Ground-truth actions are represented by \textit{foreground} frames. We then define \textit{transition} frames as those occurring within a short temporal duration just before an action begins and those occurring within a short temporal duration immediately after the action concludes. For this experiment, we set the intervals to 2 seconds. The remaining frames are considered as \textit{background}. 

\noindent\textbf{Are captions sufficient?}
We aim to determine whether textual information alone is sufficient to distinguish foreground frames from background frames. 
To this end, we first leverage $\Phi_\text{\texttt{CAP}}$ to generate captions for each frame of the videos within THUMOS14.
We then compute the similarity between the video's ground-truth class name and (i) the frame-level captions (\ie, text-to-text, using SentenceBERT) and (ii) the frame itself (\ie, text-to-image, using the VLM's encoders $\left( \phi_T, \phi_V\right)$).
Fig.~\ref{fig:sim_vlm} and Fig.~\ref{fig:sim_sbert} illustrate the two similarity scores aggregated by foreground, transition, and background, averaged per category.
Ideally, a robust discriminative model would yield high similarity scores for foreground frames, moderate scores for transition frames, and low scores for background frames, thereby achieving a clear distinction among these categories.
The figures highlight that the text-to-text similarity values derived from frame-level captions for foreground and non-foreground images are closer than the text-to-image computed from the actual visual content. 
This makes captions less effective in distinguishing foreground, transition, and background frames. 

\begin{figure*}
    \centering
    \begin{subfigure}{\linewidth}
        \centering
        \includegraphics[width=\linewidth]{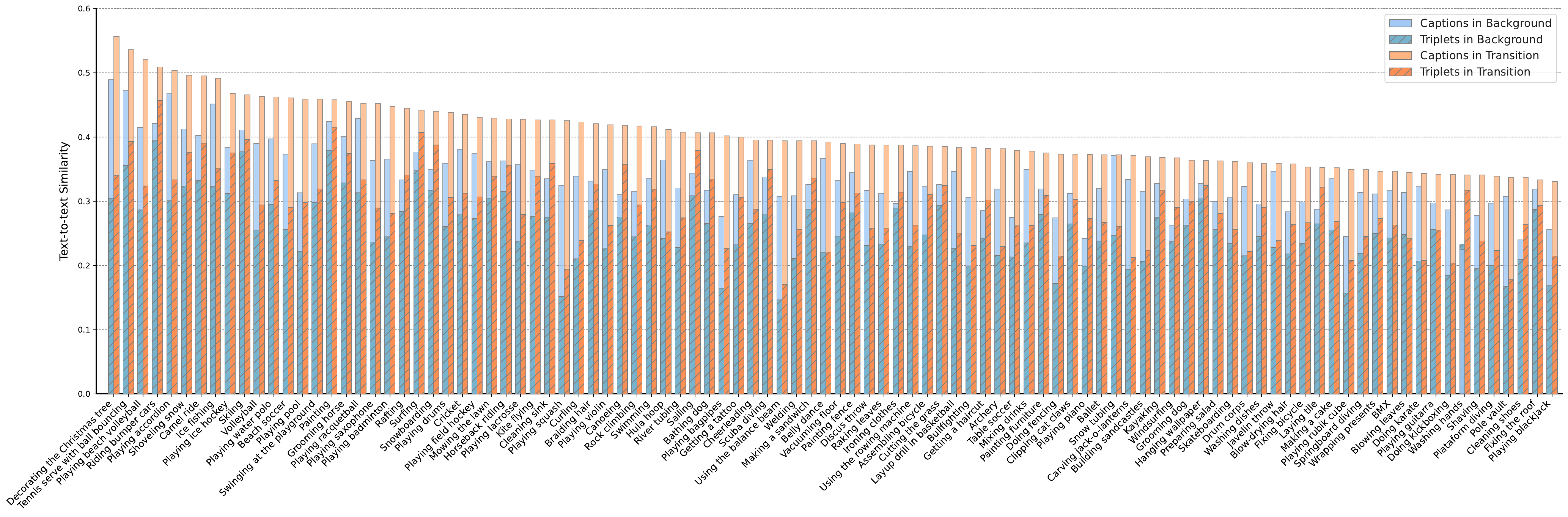}
    \end{subfigure}
    \caption{\textbf{Text-to-text similarities in non-foreground frames.} Cosine similarities between ground truth video classes and (i) frame-level captions grouped by \inlineColorbox{DrawioBlue}{background} and \inlineColorbox{DrawioOrange}{transition} and (ii) scene triplets, again grouped by \inlineColorbox{DrawioBrightBlue}{background} and \inlineColorbox{DrawioBrightOrange}{transition}. Values are averaged per-class on ActivityNet-v1.3.
    \vspace{-10pt}}
    \label{fig:anet_triplets}
\end{figure*}

\begin{figure}
    \centering
    \includegraphics[width=1.\linewidth]{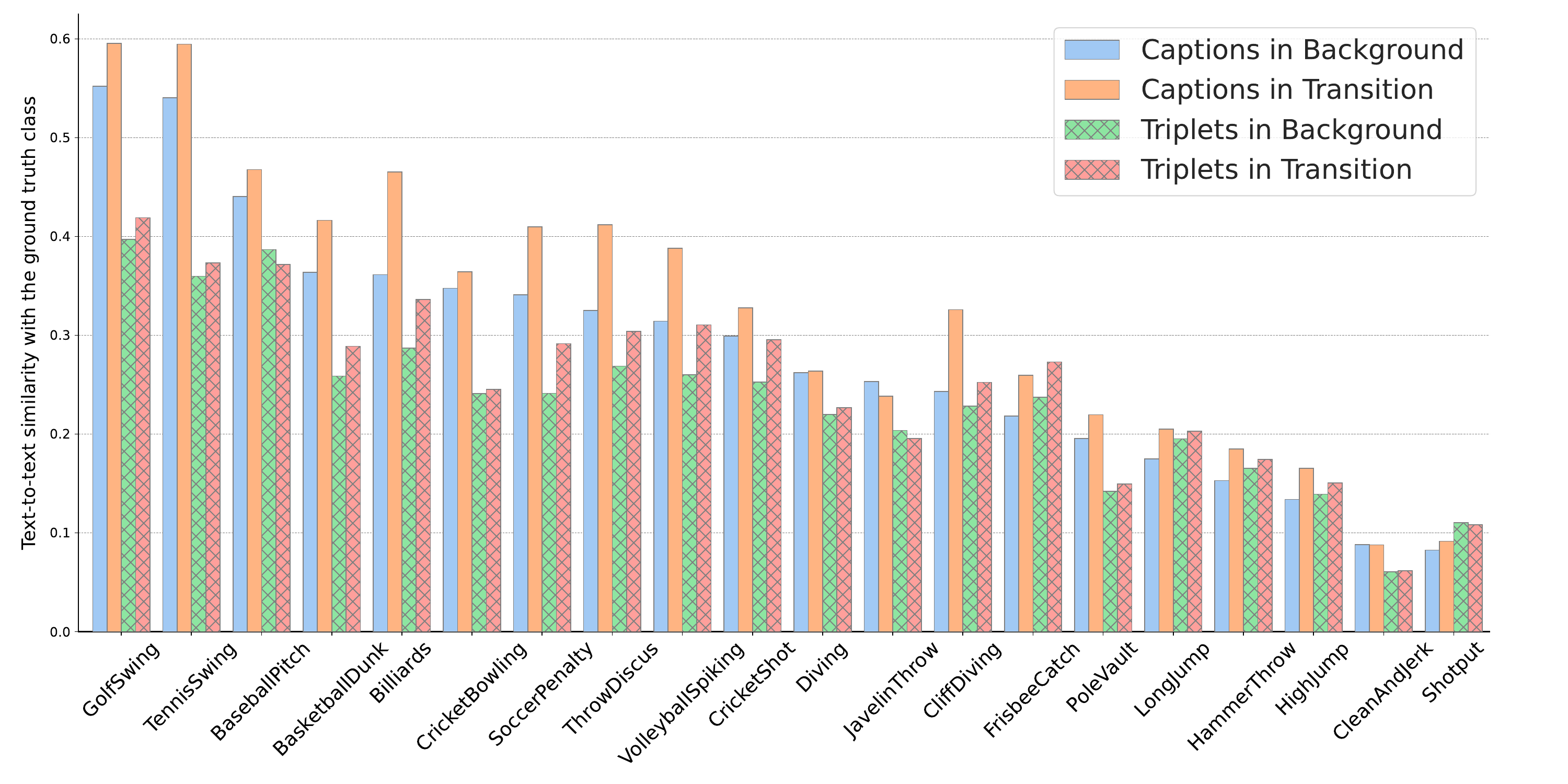}
    \caption{\textbf{Text-to-text similarities in non-foreground frames.} Cosine similarities between ground truth video classes and (i) frame-level captions grouped by \inlineColorbox{DrawioBlue}{background} and \inlineColorbox{DrawioOrange}{transition} and (ii) scene triplets, again grouped by \inlineColorbox{DrawioGreen}{background} and \inlineColorbox{DrawioRed}{transition}. Values are averaged per-class.
    \vspace{-20pt}}
\label{fig:bar_plot_triplets}
\end{figure}

\noindent \textbf{Captions are ambiguous.}
We posit that the limited discriminant power of captions is closely tied to the inherent ambiguity frequently found in human language, both spoken and written.
This ambiguity diminishes their ability to distinguish between foreground, transition, and background frames in TAL.
This may stem from the use of words that imply uncertainty regarding actions, such as \textit{``likely''}, \textit{``probably''} or \textit{``preparing to''}, resulting in increased uncertainty about whether the action is actually performed.
As a result, the intended meaning of captions is diluted, leading to a higher likelihood of false positives. 
To quantify the level of action ambiguity in our analysis, we compute the percentage of captions that contain at least one of these ambiguous terms. 
Our findings on THUMOS14 indicate that $21.2\%$ of generated captions contain such ambiguous terms.
We suggest that this ambiguity also contributes to the observed phenomenon where captions correlate highly with class labels, even in non-action frames.

\noindent\textbf{Can scene triplets help ZS-TAL?}
Using captions directly is often not advantageous for ZS-TAL and may even hinder performance, largely due to their limited discriminative power and ambiguity.
To address this, we design an additional experiment to evaluate the benefits of deriving structured triplets from free-form captions.
For each video $\mathcal{V}$, we obtain scene triplets $\mathcal{T}_V$ and cluster them as detailed in Sec~\ref{sec:method}, obtaining $\mathcal{T}_V^S$. 
Then, for each frame $x_i \in \mathcal{V}$ we replace its triplets $\mathcal{T}_V$ with the closest centroid in $\mathcal{T}_V^S$. 
Similar to the previous analysis, we compute the cosine similarity of the updated scene triplets against the ground truth video class, by computing uni-modal cosine similarity with SentenceBERT embeddings.
As shown in Fig.~\ref{fig:bar_plot_triplets}, triplets exhibit lower similarity scores in background and transition regions, compared to the captions. In particular, we see a significant decrease in similarity values for very atomic actions (\eg, ``\textit{tennis swing}") that are particularly challenging for precise ZS-TAL due to the high similarity between frames in the foreground and transition regions. 
A consistent trend is also observed on ActivityNet-v1.3, as shown in Fig.~\ref{fig:anet_triplets}.
Additional results for this dataset are provided in the \suppmat.

%% file: tables/thumos.tex
\begin{table}[t]
\small
\centering
\caption{\textbf{Results on THUMOS14.} Comparison of \inlineColorbox{OurMethodColor}{our method} with state-of-the-art \inlineColorbox{white}{training-based} and \inlineColorbox{DrawioOrange}{non-training-based} approaches. }
\resizebox{\columnwidth}{!}{
\begin{tabular}{llccccc|c}
\toprule
\textbf{Setting} & \textbf{Method} & \multicolumn{6}{c}{\textbf{mAP (\%) $\uparrow$}} \\
           &      & \textbf{0.3} & \textbf{0.4} & \textbf{0.5} &     \textbf{0.6} & \textbf{0.7} & \textbf{Avg.}  \\ 
\midrule

\multirow{6}{*}{$75\%$-$25\%$} & \cellcolor{white}Detector + CLIP~\cite{Ju2021PromptingVM}  &\cellcolor{white} 33.0 &\cellcolor{white} 25.5 &\cellcolor{white} 18.3 &\cellcolor{white} 11.6 &\cellcolor{white} 5.7 &\cellcolor{white} 18.8 \\
&\cellcolor{white} EffPrompt~\cite{Ju2021PromptingVM}     & \cellcolor{white}39.7 & \cellcolor{white}31.6 & \cellcolor{white}23.0 & \cellcolor{white}14.9 & \cellcolor{white}7.5 & \cellcolor{white}23.3 \\
 & \cellcolor{white} STALE~\cite{stale}  &\cellcolor{white} 40.5 &\cellcolor{white} 32.3 &\cellcolor{white} 23.5 &\cellcolor{white} 15.3 &\cellcolor{white} 7.6 &\cellcolor{white} 23.8  \\
& \cellcolor{white} ZEETAD~\cite{phan2024zeetad} &\cellcolor{white} 61.4 & \cellcolor{white}53.9 & \cellcolor{white}44.7 & \cellcolor{white}34.5 &\cellcolor{white} 20.5 &\cellcolor{white} 43.2 \\
 
\cmidrule(lr){2-8}
& \cellcolor{DrawioOrange} $T3AL$~\cite{Liberatori2024}                          & \cellcolor{DrawioOrange} 19.2 & \cellcolor{DrawioOrange} 12.7 & \cellcolor{DrawioOrange} 7.4 & \cellcolor{DrawioOrange} 4.4 & \cellcolor{DrawioOrange} 2.2 & \cellcolor{DrawioOrange} 9.2 \\

& \cellcolor{OurMethodColor}\methodname & \cellcolor{OurMethodColor}23.3 & \cellcolor{OurMethodColor}15.7 &\cellcolor{OurMethodColor} 9.1 &\cellcolor{OurMethodColor} 4.7 & \cellcolor{OurMethodColor}2.4 &\cellcolor{OurMethodColor} 11.0              \\
\bottomrule

\multirow{6}{*}{$50\%$-$50\%$} & \cellcolor{white}Detector + CLIP~\cite{Ju2021PromptingVM}     & \cellcolor{white} 27.2 & \cellcolor{white} 21.3 & \cellcolor{white} 15.3 &\cellcolor{white}  9.7 &\cellcolor{white} 4.8 &\cellcolor{white} 15.7  \\

& \cellcolor{white} EffPrompt~\cite{Ju2021PromptingVM}     &\cellcolor{white} 37.2 &\cellcolor{white} 29.6 &\cellcolor{white} 21.6 & \cellcolor{white} 14.0 &\cellcolor{white} 7.2 & \cellcolor{white}21.9 \\

 &\cellcolor{white} STALE~\cite{stale} &\cellcolor{white} 38.3 &\cellcolor{white} 30.7 &\cellcolor{white} 21.2 &\cellcolor{white} 13.8 & \cellcolor{white} 7.0 & \cellcolor{white}22.2 \\
 &\cellcolor{white} ZEETAD~\cite{phan2024zeetad} &\cellcolor{white} 45.2 &\cellcolor{white} 38.8 &\cellcolor{white} 30.8 &\cellcolor{white} 22.5 &\cellcolor{white} 13.7 &\cellcolor{white} 30.2 \\
\cmidrule(lr){2-8}
& \cellcolor{DrawioOrange} $T3AL$~\cite{Liberatori2024}  & \cellcolor{DrawioOrange} 20.7 & \cellcolor{DrawioOrange} 14.3 & \cellcolor{DrawioOrange} 8.9 & \cellcolor{DrawioOrange} 5.3 &\cellcolor{DrawioOrange} 2.7 & \cellcolor{DrawioOrange} 10.4 \\

& \cellcolor{OurMethodColor} \methodname & \cellcolor{OurMethodColor}22.4 &\cellcolor{OurMethodColor} 14.8 & \cellcolor{OurMethodColor} 8.6 &\cellcolor{OurMethodColor} 4.3	&\cellcolor{OurMethodColor} 2.2 &\cellcolor{OurMethodColor} 10.5 \\

\bottomrule
\end{tabular}
}
\label{thumos}
\end{table}

%% file: tables/anet.tex
\begin{table}[t]
\small
\centering
\caption{\textbf{Results on ActivityNet-v1.3.}   Comparison of \inlineColorbox{OurMethodColor}{our method} with state-of-the-art \inlineColorbox{white}{training-based} and \inlineColorbox{DrawioOrange}{non-training-based} approaches.}
\resizebox{0.9\columnwidth}{!}{
\begin{tabular}{llccc|c}
\toprule
\textbf{Setting} & \textbf{Method} & \multicolumn{4}{c}{\textbf{mAP (\%) $\uparrow$}} \\
        &        & \textbf{0.50} &  \textbf{0.75} & \textbf{0.95} & \textbf{Avg.}  \\ 
\midrule

\multirow{7}{*}{$75\%$-$25\%$} &\cellcolor{white}Detector + CLIP~\cite{Ju2021PromptingVM} & \cellcolor{white}35.6 &\cellcolor{white}  20.4 & \cellcolor{white}2.1 &\cellcolor{white} 20.2 \\

& \cellcolor{white}EffPrompt~\cite{Ju2021PromptingVM}     & \cellcolor{white}37.6 &\cellcolor{white} 22.9 & \cellcolor{white}3.8 & \cellcolor{white}23.1 \\

& \cellcolor{white}STALE~\cite{stale}                     &\cellcolor{white} 38.2 &\cellcolor{white} 25.2 & \cellcolor{white}6.0 &\cellcolor{white} 24.9 \\

& \cellcolor{white}UnLoc~\cite{Yan_2023_ICCV}             & \cellcolor{white} 48.8 & \cellcolor{white}-  & \cellcolor{white}- & \cellcolor{white}-  \\
&\cellcolor{white} ZEETAD~\cite{phan2024zeetad} & \cellcolor{white} 51.0 & \cellcolor{white} 33.4 & \cellcolor{white} 5.9 & \cellcolor{white} 32.5 \\
\cmidrule(lr){2-6}
& \cellcolor{DrawioOrange} $T3AL$~\cite{Liberatori2024}  & \cellcolor{DrawioOrange} 28.1 &	\cellcolor{DrawioOrange} 14.9 & \cellcolor{DrawioOrange} 3.3 & \cellcolor{DrawioOrange}	15.4 \\

&\cellcolor{OurMethodColor} \methodname  & \cellcolor{OurMethodColor} 34.7	& \cellcolor{OurMethodColor} 19.6	& \cellcolor{OurMethodColor} 2.8	& \cellcolor{OurMethodColor} 19.0 \\
\bottomrule
\multirow{7}{*}{$50\%$-$50\%$} &\cellcolor{white} Detector + CLIP~\cite{Ju2021PromptingVM}  & \cellcolor{white}28.0 & \cellcolor{white} 16.4 &\cellcolor{white} 1.2 &\cellcolor{white} 16.0 \\
& \cellcolor{white} EffPrompt~\cite{Ju2021PromptingVM} & \cellcolor{white} 32.0 & \cellcolor{white}  19.3 & \cellcolor{white}  2.9 & \cellcolor{white}  19.6 \\
 & \cellcolor{white} STALE~\cite{stale}     & \cellcolor{white} 32.1 &\cellcolor{white} 20.7 &\cellcolor{white} 5.9 &\cellcolor{white} 20.5 \\
& \cellcolor{white} UnLoc~\cite{Yan_2023_ICCV}             & \cellcolor{white} 43.7 & \cellcolor{white}-  & \cellcolor{white}-  & \cellcolor{white}-  \\
&\cellcolor{white} ZEETAD~\cite{phan2024zeetad} & \cellcolor{white} 39.2 & \cellcolor{white} 25.7 & \cellcolor{white} 3.1 & \cellcolor{white} 24.9 \\
\cmidrule(lr){2-6}

& \cellcolor{DrawioOrange} $T3AL$~\cite{Liberatori2024} & \cellcolor{DrawioOrange} 25.8 & \cellcolor{DrawioOrange} 13.9 & \cellcolor{DrawioOrange} 3.1 & \cellcolor{DrawioOrange} 14.3 \\

 & \cellcolor{OurMethodColor}\methodname  & \cellcolor{OurMethodColor} 33.5 & \cellcolor{OurMethodColor} 19.3 & \cellcolor{OurMethodColor} 2.6	& \cellcolor{OurMethodColor} 18.5 \\
\bottomrule
\end{tabular}
}
\label{anet}
\end{table}

%% file: figures/clustering.tex
\begin{figure}
    \centering
    \includegraphics[width=.9\linewidth]{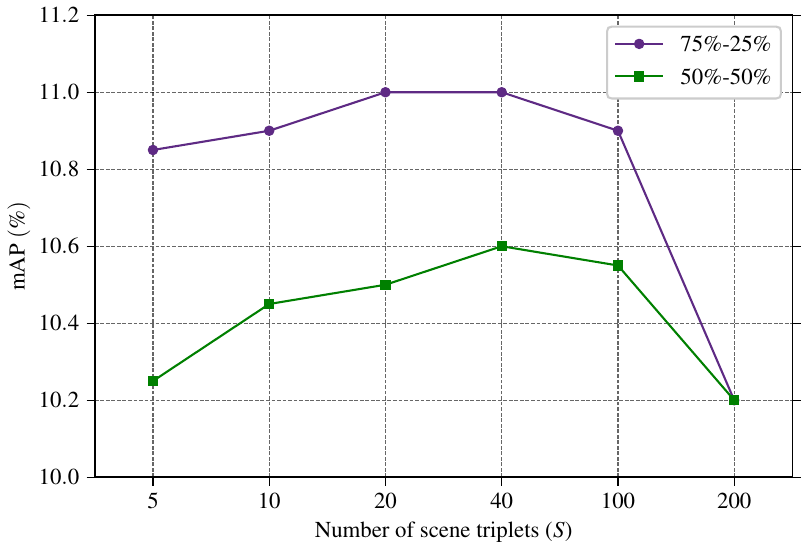}
    \caption{\textbf{Impact of varying the number of $S$ scene triplets.} Results are collected on THUMOS14 \inlineColorbox{DrawioGreen}{($50\%$-$50\%$)} and on THUMOS14 \inlineColorbox{DrawioPurple}{($75\%$-$25\%$)}.}
    \label{fig:kmeans}
\end{figure}

%% file: tables/ablation0.tex
\begin{table}
\small
\centering
\caption{\textbf{Ablation on the use of descriptions $\mathcal{D}$ and triplets $\mathcal{T}$.} Results are collected on THUMOS14 ($50\%$-$50\%$) and on THUMOS14 ($75\%$-$25\%$). \methodname \inlineColorbox{OurMethodColor}{final configuration} is highlighted. }
\resizebox{.9\columnwidth}{!}{
\begin{tabular}{lll|ccccc|c}
\toprule
\textbf{Setting}  & \textbf{$\mathcal{D}$} & \textbf{$\mathcal{T}$} & \multicolumn{6}{c}{\textsc{\textbf{mAP (\%) $\uparrow$}}} \\
                       & &       & \textbf{0.3} & \textbf{0.4} & \textbf{0.5} & \textbf{0.6} & \textbf{0.7} & \textbf{Avg.}  \\ 
\midrule
\multirow{4}{*}{$50\%$-$50\%$} & &  &              16.9 & 10.7	& 6.1	& 3.2	& 1.6	& 7.7\\
 & & \checkmark &    17.3 &	11.7	& 6.9	& 3.6	& 1.8	& 8.3   \\
 & \checkmark & &    22.1 &  14.6 &	8.2	 & 4.1	& 2.1	& 10.2  \\
 & \checkmark & \checkmark & \cellcolor{OurMethodColor}22.4 & \cellcolor{OurMethodColor}14.8 & \cellcolor{OurMethodColor}8.6 & \cellcolor{OurMethodColor}4.3 & \cellcolor{OurMethodColor}2.2 & \cellcolor{OurMethodColor}10.5 \\
\midrule
\multirow{4}{*}{$75\%$-$25\%$} &  &                     &  17.6	& 11.5	& 6.7 & 3.5	 & 1.8	& 8.2     \\
 &  & \checkmark  &  17.8	& 12.2	& 7.5 & 3.8	 & 1.9	& 8.6     \\
 & \checkmark &  &         23.1	& 15.5	& 8.7	& 4.3	& 2.3	& 10.8 \\
 & \checkmark &\checkmark & \cellcolor{OurMethodColor}23.3 & \cellcolor{OurMethodColor}15.7 & \cellcolor{OurMethodColor}9.1 & \cellcolor{OurMethodColor}4.7 & \cellcolor{OurMethodColor}2.4 & \cellcolor{OurMethodColor}11.0 \\ 
\bottomrule
\end{tabular}
}
\label{table:text}
\end{table}

%% file: tables/ablation2.tex
\begin{table}[t]
\small
\centering
\caption{\textbf{Ablation on model adaptation and learning objective.} Results are collected on THUMOS14 ($50\%$-$50\%$) and on THUMOS14 ($75\%$-$25\%$). \methodname \inlineColorbox{OurMethodColor}{final configuration} is highlighted. }
\resizebox{.9\columnwidth}{!}{
\begin{tabular}{ll|ccccc|c}
\toprule
\textbf{Setting} & \textbf{Loss} & \multicolumn{6}{c}{\textbf{mAP (\%) $\uparrow$}} \\
                      &      & \textsc{\textbf{0.3}} & \textbf{0.4} & \textbf{0.5} & \textbf{0.6} & \textbf{0.7} & \textbf{Avg.}  \\ 
\midrule
\multirow{3}{*}{$50\%$-$50\%$} & \quad\ding{55}  & 16.6 & 10.7 & 6.2 & 3.1	& 1.4 & 7.6  \\
& \textsc{BYOL} & 20.1 & 13.5 & 8.0 &	4.1	& 2.1 & 9.6 \\
&  $\mathcal{L}_m$ &\cellcolor{OurMethodColor} 22.4 & \cellcolor{OurMethodColor} 14.8	& \cellcolor{OurMethodColor} 8.6 & \cellcolor{OurMethodColor} 4.3	&\cellcolor{OurMethodColor} 2.2 & \cellcolor{OurMethodColor} 10.5 \\
\midrule
\multirow{3}{*}{$75\%$-$25\%$}& \quad\ding{55}  & 16.8 & 10.8 & 6.4 & 3.1	& 1.4 & 7.7 \\
& \textsc{BYOL} & 20.2 &	13.5 & 8.1	& 4.1	& 2.1	& 9.6 \\
& $\mathcal{L}_m$  & \cellcolor{OurMethodColor} 23.3 &\cellcolor{OurMethodColor} 15.7 &\cellcolor{OurMethodColor} 9.1 &\cellcolor{OurMethodColor} 4.7	& \cellcolor{OurMethodColor} 2.4 &\cellcolor{OurMethodColor} 11.0 \\
\bottomrule
\end{tabular}
}
\label{table:loss}
\end{table}

%% file: sec/conclusion.tex
\section{Conclusion} \label{sec:conclusion}

In this work, we explored the potential of integrating complementary textual data to improve performance in the challenging ZS-TAL task. Our findings indicate that directly applying frame captions is insufficient, as captions often contain redundant and noisy information, which may hinder accurate temporal action localization. To address this, we introduced \methodname{}, a novel approach that extracts refined semantic concepts using scene triplets from captions. This approach, coupled with a test-time adaptation framework, optimizes localization performance.

Experimental results on two benchmarks demonstrate that \methodname~surpasses state-of-the-art methods in challenging conditions, validating our design choices. Furthermore, we offer valuable insights into the advantages and limitations of caption-derived textual information, highlighting practical strategies for its effective utilization in ZS-TAL.

\noindent\textbf{Limitations.} While our method completely removes the need for human annotations, 
it suffers from some limitations. First, it 
depends on generative models, which 
bound its performance.
Finally, the performance is far from saturated, highlighting the challenges associated with this task and encouraging further exploration.

\section*{Acknowledgments}
\small This work was partially supported by the EU Horizon projects ELLIOT (No. 101214398), and IAMI (No. 101168272).
We acknowledge the Deep Learning
Lab of the ProM Facility for the GPU time.

%% file: supsec/supplementary.tex
\clearpage
\setcounter{page}{1}
\setcounter{figure}{8}  
\setcounter{table}{4}
\setcounter{section}{0}

\noindent In this Supplementary Material, we provide additional quantitative and qualitative results.
In Sec.~\ref{sec:classwise}, we report additional results; in Sec.~\ref{sec:additionalanalyses}, we extend the analysis presented in the main paper regarding captions and scene triplets.
Following that, in Sec.~\ref{sec:prompts}, we provide the exact form of the prompts employed in the main paper; in Sec.~\ref{sec:impl}, we outline the remaining implementation details.

\input{supsec/classwise}

\input{supsec/captions}

\input{supsec/prompts}
\input{supsec/details}
\input{supsec/impact}

%% file: supsec/classwise.tex
\section{Additional results}\label{sec:classwise}
We report per-class performance of \methodname~on the THUMOS14 dataset~\cite{IDREES20171} for both evaluation settings, \ie, $50\%$-$50\%$ in \cref{tab:supp_50} and $75\%$-$25\%$ in \cref{tab:supp_75}.
Notably, the $75\%$-$25\%$ setup excludes two classes because these are absent from all test splits in this configuration.
Consistent with~\cite{Ju2021PromptingVM}, the reported results represent the averages of the individual results obtained across all class splits. 
Both tables reveal significant variability in performance across different classes, highlighting the challenges posed by class-specific disparities.
Specifically, actions with clear and distinctive visual cues, such as ``\textit{CliffDiving}'' and ``\textit{PoleVault}'', achieve consistently higher scores.
These actions are easier to recognize because their key features, \eg, the body posture or equipment used, are visually unique and easy to distinguish from other actions.
In contrast, actions like ``\textit{TennisSwing}'' or ``\textit{CricketShot}'' present a greater challenge as they require a more precise identification of the action's foreground, such as the ball's impact or the swing's apex.
These moments occur within sequences that may otherwise appear visually similar, making it harder for the model to distinguish them accurately.
\input{suptables/classwise50}
\input{suptables/classwise75}

In terms of classification performance, we achieved Top1/5 Accuracy of $79.37/99.27\%$  on THUMOS14 and $82.56/96.68\%$ on ActivityNet-v1.3.
Moreover, when using the ground truth action classes, we obtain a $+0.95/1.59\%$ and $+1.40/2.36\%$ in the average mAP for both settings and datasets.
These results indicate that precise boundary localization is the most challenging aspect of TAL, rather than matching the classes of the predicted boundaries.

Lastly, in Tab.~\ref{table:supp_backbone} we extend the results of the main paper to compare \methodname to prior work~\cite{Liberatori2024}, evaluated under the same VLM backbone configuration. 
Specifically, we use SigLIP~\cite{zhai2023sigmoid} and CoCa~\cite{yu2022coca} for both \methodname and $T3AL$~\cite{Liberatori2024}. 
In comparison with the same backbone, our results show that the proposed text-guided adaptation strategy is more effective, highlighting the advantages of our overall approach.

\input{suptables/backbone}

%% file: suptables/classwise50.tex
\begin{table}[ht]
    \centering
    \resizebox{\columnwidth}{!}{
    \begin{tabular}{lccccc|c}
        \toprule
        \textbf{Class Name} & \multicolumn{6}{c}{\textbf{mAP (\%) $\uparrow$}} \\
                & \textbf{0.3} & \textbf{0.4} & \textbf{0.5} & \textbf{0.6} & \textbf{0.7} & \textbf{Avg.}  \\ 
\midrule
 BaseballPitch     & 18.3  & 12.7  &  5.3  &  2.6  &  1.6  &  8.1   \\
 Billiards         &  8.1  &  4.2  &  1.4  &  0.4  &  0.1  &  2.8   \\
 CleanAndJerk      & 37.9  & 23.5  & 14.5  &  10.0   &  5.8  &  18.3  \\
 GolfSwing         & 29.1  & 23.2  & 14.9  &  2.4  &  0.9  &  14.1  \\
 HammerThrow       & 16.9  & 14.3  &  10   &  6.6  &  4.1  &  10.4  \\
 HighJump          & 45.9  & 27.9  & 14.1  &  6.8  &  2.5  &  19.4  \\
 PoleVault         & 47.2  & 32.8  & 20.7  & 11.2  &   6.0   &  23.6  \\
 SoccerPenalty     & 26.2  &  17.0   &  9.4  &   4.0   &  0.8  &  11.5  \\
 TennisSwing       &  7.2  &   4.0   &  1.7  &  0.4  &  0.1  &  2.7   \\
 ThrowDiscus       & 13.9  &  9.8  &  5.7  &  3.4  &   2.0   &   7.0    \\
 BasketballDunk    & 31.5  & 21.4  &  13.0   &  6.3  &  2.7  &   15.0   \\
 CricketBowling    &  6.1  &  3.2  &  1.7  &  0.6  &  0.4  &  2.4   \\
 CricketShot       &  2.2  &   1.0   &  0.3  &  0.1  &   0.0   &  0.7   \\
 Shotput           & 14.2  &  9.8  &  5.5  &  4.1  &  2.9  &  7.3   \\
 VolleyballSpiking &  23   & 15.9  &  8.3  &  3.5  &  1.7  &  10.5  \\
 Diving            & 34.5  & 24.1  & 13.9  &  7.9  &   4.0   &  16.9  \\
 FrisbeeCatch      & 17.9  &  8.7  &  4.3  &  2.2  &  1.2  &  6.9   \\
 LongJump          & 21.9  & 13.6  &  7.3  &   3.0   &  1.4  &  9.4   \\
 JavelinThrow      & 30.6  & 19.8  & 11.9  &  7.1  &  3.2  &  14.5  \\
 CliffDiving       & 37.8  & 28.7  & 18.1  & 11.2  &   7.0   &  20.5  \\
\bottomrule
    \end{tabular}
    }
    \caption{\textbf{Class-wise results on THUMOS14 ($50\%$-$50\%$).} Numbers are computed at IoU thresholds of [$0.3$:$0.1$:$0.7$] and averaged across all class splits.}
    \label{tab:supp_50}
\end{table}

%% file: suptables/classwise75.tex
\begin{table}[ht]
    \centering
    \resizebox{\columnwidth}{!}{
    \begin{tabular}{lccccc|c}
        \toprule
        \textbf{Class Name} & \multicolumn{6}{c}{\textbf{mAP (\%) $\uparrow$}} \\
                & \textbf{0.3} & \textbf{0.4} & \textbf{0.5} & \textbf{0.6} & \textbf{0.7} & \textbf{Avg.} \\
\midrule
 BaseballPitch     & 23.8  & 16.6  &  7.4  &  3.3  &  2.1  &  10.7  \\
 HammerThrow       &  0.8  &  0.4  &  0.2  &  0.1  &   0.0   &  0.3   \\
 PoleVault         & 51.3  & 35.9  & 23.2  & 12.5  &  6.9  &   26.0   \\
 SoccerPenalty     & 25.1  & 15.1  &  7.7  &  3.2  &  0.8  &  10.4  \\
 ThrowDiscus       & 11.9  &  8.8  &  5.3  &  3.2  &   2.0   &  6.2   \\
 CricketBowling    &  6.5  &  3.6  &  1.6  &  0.5  &  0.4  &  2.5   \\
 GolfSwing         & 28.8  & 25.2  &  15.0   &  2.1  &  0.9  &  14.4  \\
 HighJump          & 53.5  & 33.1  &  15.0   &  6.9  &  2.5  &  22.2  \\
 TennisSwing       &  7.2  &  4.1  &  1.7  &  0.5  &  0.1  &  2.7   \\
 VolleyballSpiking & 22.4  & 15.6  &  8.1  &  3.7  &  1.8  &  10.3  \\
 CricketShot       &  2.7  &  1.2  &  0.5  &  0.1  &  0.1  &  0.9   \\
 Diving            & 35.6  & 24.9  & 14.5  &  8.6  &  3.9  &  17.5  \\
 FrisbeeCatch      & 14.1  &  6.8  &  3.7  &   2.0   &  0.8  &  5.5   \\
 CliffDiving       & 38.9  & 28.3  & 18.1  & 11.7  &  7.3  &  20.9  \\
 JavelinThrow      & 36.5  & 24.1  & 14.6  &  8.4  &  4.1  &  17.5  \\
 Shotput           &  3.1  &  1.5  &  0.7  &  0.7  &  0.5  &  1.3   \\
\bottomrule
    \end{tabular}
    }
    \caption{\textbf{Class-wise results on THUMOS14 ($75\%$-$25\%$).} Numbers are computed at IoU thresholds of [$0.3$:$0.1$:$0.7$] and averaged across all class splits.}
    \label{tab:supp_75}
\end{table}

%% file: suptables/backbone.tex
\begin{table}[t]
\small
\centering
\resizebox{1.0\columnwidth}{!}{
\begin{tabular}{@{}lll|ccccc|c@{}}
\toprule
\textbf{Setting} & \textbf{Method} & \textbf{Backbone} & \multicolumn{6}{c}{\textbf{mAP (\%) $\uparrow$}} \\
& & & \textbf{0.3} & \textbf{0.4} & \textbf{0.5} & \textbf{0.6} & \textbf{0.7} & \textbf{Avg.} \\ 
\midrule
\multirow{2}{*}{$50\%$-$50\%$} & $T3AL$~\cite{Liberatori2024} & CoCa~\cite{yu2022coca} &  20.7 & 14.3 & 8.9 & 5.3 & 2.7 & 10.4\\
& $T3AL$~\cite{Liberatori2024} & SigLIP~\cite{zhai2023sigmoid}   & 19.2	   & 11.9  & 7.0	 & 4.1    &	2.1	  &   8.9  \\
& \methodname & CoCa~\cite{yu2022coca} & 20.9 & 14.1 & 8.8 & 5.2 & 2.6 & 10.3 \\
& \methodname & SigLIP~\cite{zhai2023sigmoid}  & \cellcolor{OurMethodColor}23.3 & \cellcolor{OurMethodColor}15.7 & \cellcolor{OurMethodColor}9.1 & \cellcolor{OurMethodColor}4.7 & \cellcolor{OurMethodColor}2.4 & \cellcolor{OurMethodColor}11.0 \\
\midrule
\multirow{2}{*}{$75\%$-$25\%$}& $T3AL$~\cite{Liberatori2024}  & CoCa~\cite{yu2022coca} & 19.2 & 12.7 & 7.4 & 4.4 & 2.2 & 9.2 \\
& $T3AL$~\cite{Liberatori2024} & SigLIP~\cite{zhai2023sigmoid} & 19.4	& 12.4	& 7.5 & 4.4   &	2.3	  &   9.2  \\
& \methodname & CoCa~\cite{yu2022coca} & 20.5 & 13.2 & 7.9 &  4.5 & 2.3 & 9.7 \\
& \methodname & SigLIP~\cite{zhai2023sigmoid}  & \cellcolor{OurMethodColor}22.4 & \cellcolor{OurMethodColor}14.8 & \cellcolor{OurMethodColor}8.6 & \cellcolor{OurMethodColor}4.3 & \cellcolor{OurMethodColor}2.2 & \cellcolor{OurMethodColor}10.5 \\
\bottomrule
\end{tabular}
}
\caption{\textbf{Performance of \methodname and previous approaches under the same backbone setting} on THUMOS14. \methodname \inlineColorbox{OurMethodColor}{final configuration} is highlighted.}
\label{table:supp_backbone}
\end{table}

%% file: supsec/captions.tex
\section{Additional analyses on captions}\label{sec:additionalanalyses}

In this section, we build upon the analysis presented in Sec.~\refbase{4.3} for the THUMOS14 dataset~\cite{IDREES20171} by incorporating results from the ActivityNet-v1.3 dataset~\cite{7298698}. 
Additionally, we examine the effect of using different captioning models, extending the evaluation outlined in the main paper.

\begin{figure*}
    \centering
    \begin{subfigure}[t]{\linewidth}
        \centering
        \includegraphics[width=1.\linewidth]{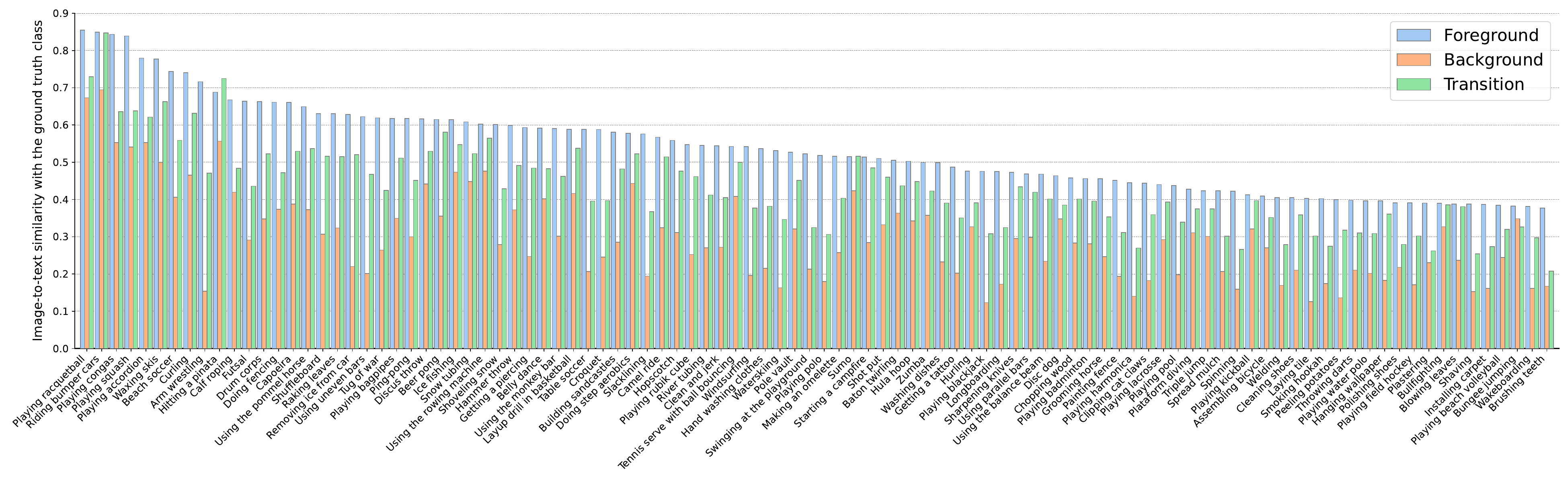}
    \end{subfigure}
    \vspace{0.5cm}
    \begin{subfigure}[t]{\linewidth}
        \centering
        \includegraphics[width=1.\linewidth]{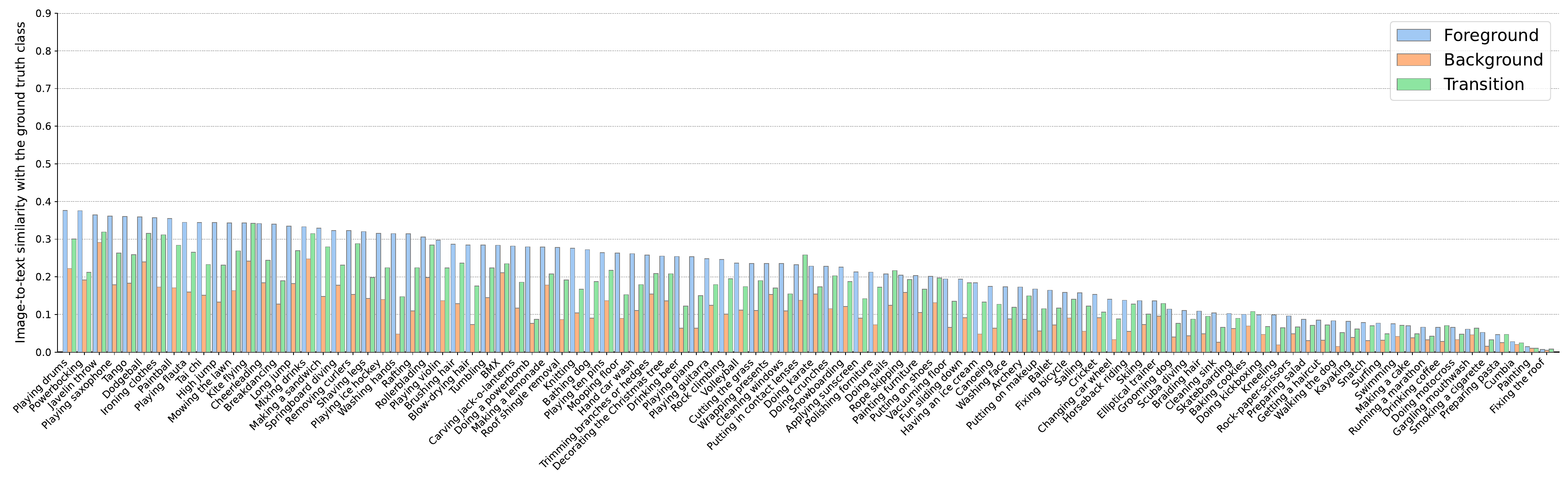}
    \end{subfigure}
    \caption{\textbf{Image-to-text similarities on ActivityNet-v1.3.} Cosine similarities between frames and ground truth video classes, computed on VLM embeddings. 
    Numbers are calculated for \inlineColorbox{DrawioBlue}{foreground}, \inlineColorbox{DrawioOrange}{background}, and \inlineColorbox{DrawioGreen}{transition} frames, and averaged across each class.}
    \label{fig:anet_vlm}
\end{figure*}

\begin{figure*}
    \centering
    \begin{subfigure}[t]{\linewidth}
        \centering
        \includegraphics[width=1.\linewidth]{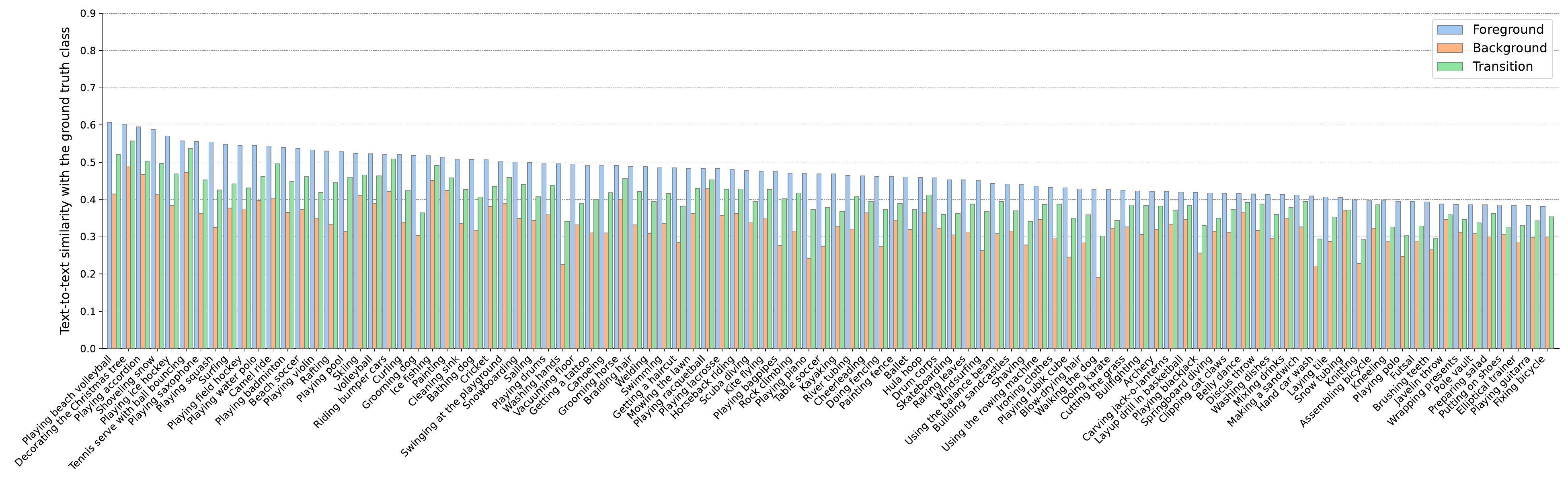}
    \end{subfigure}
    \vspace{0.5cm}
    \begin{subfigure}[t]{\linewidth}
        \centering
        \includegraphics[width=1.\linewidth]{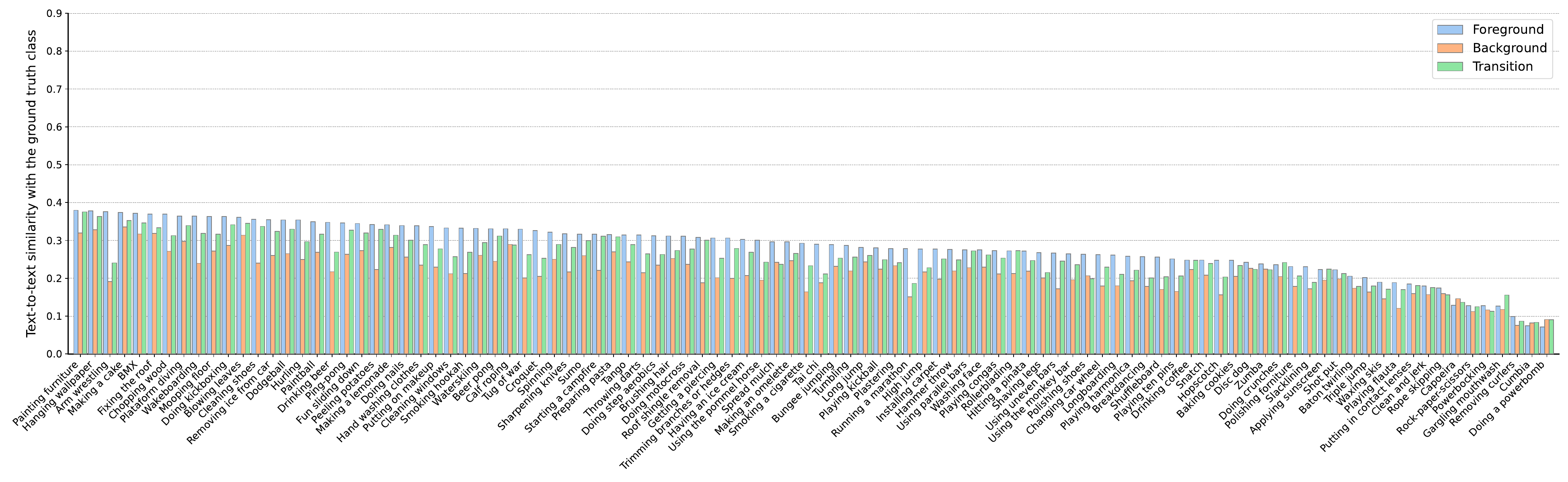}
    \end{subfigure}
    \caption{\textbf{Text-to-text similarities on ActivityNet-v1.3.} Cosine similarities between frame-level captions and ground truth video classes, computed on SentenceBERT embeddings. 
    Numbers are calculated for \inlineColorbox{DrawioBlue}{foreground}, \inlineColorbox{DrawioOrange}{background}, and \inlineColorbox{DrawioGreen}{transition} frames, and averaged across each class.}
    \label{fig:anet_sbert}
\end{figure*}

\noindent \textbf{Captions and scene triplets on ActivityNet-v1.3.} 
As shown in the main paper with the THUMOS14 dataset in Figs.~\refbase{5} and~\refbase{6},
\cref{fig:anet_vlm} displays the \emph{image-to-text} similarities \emph{between video frames and their corresponding ground-truth video classes}.
These scores are grouped into foreground, transition, and background categories and then averaged across each class.
Similarly, \cref{fig:anet_sbert} illustrates the \emph{text-to-text} similarity scores, calculated \emph{between frame-level captions and ground-truth} video classes, using the same grouping and averaging approach.
For both, we plot the results for all 200 classes in ActivityNet-v1.3, divided into two subplots for clarity.  
From this analysis, we can conclude the same trends observed in the THUMOS14 dataset, \ie, the text-to-text similarity values derived from frame-level captions for foreground and non-foreground images (\cref{fig:anet_sbert}) are closer to each other than the corresponding image-to-text similarity values (\cref{fig:anet_vlm}).
This suggests that caption-based representations lack the granularity required to align closely with the semantic information of each frame.
Consequently, they struggle to clearly distinguish the moments where the action occurs from the rest of the video content.

\begin{figure}
    \centering
    \includegraphics[width=\linewidth]{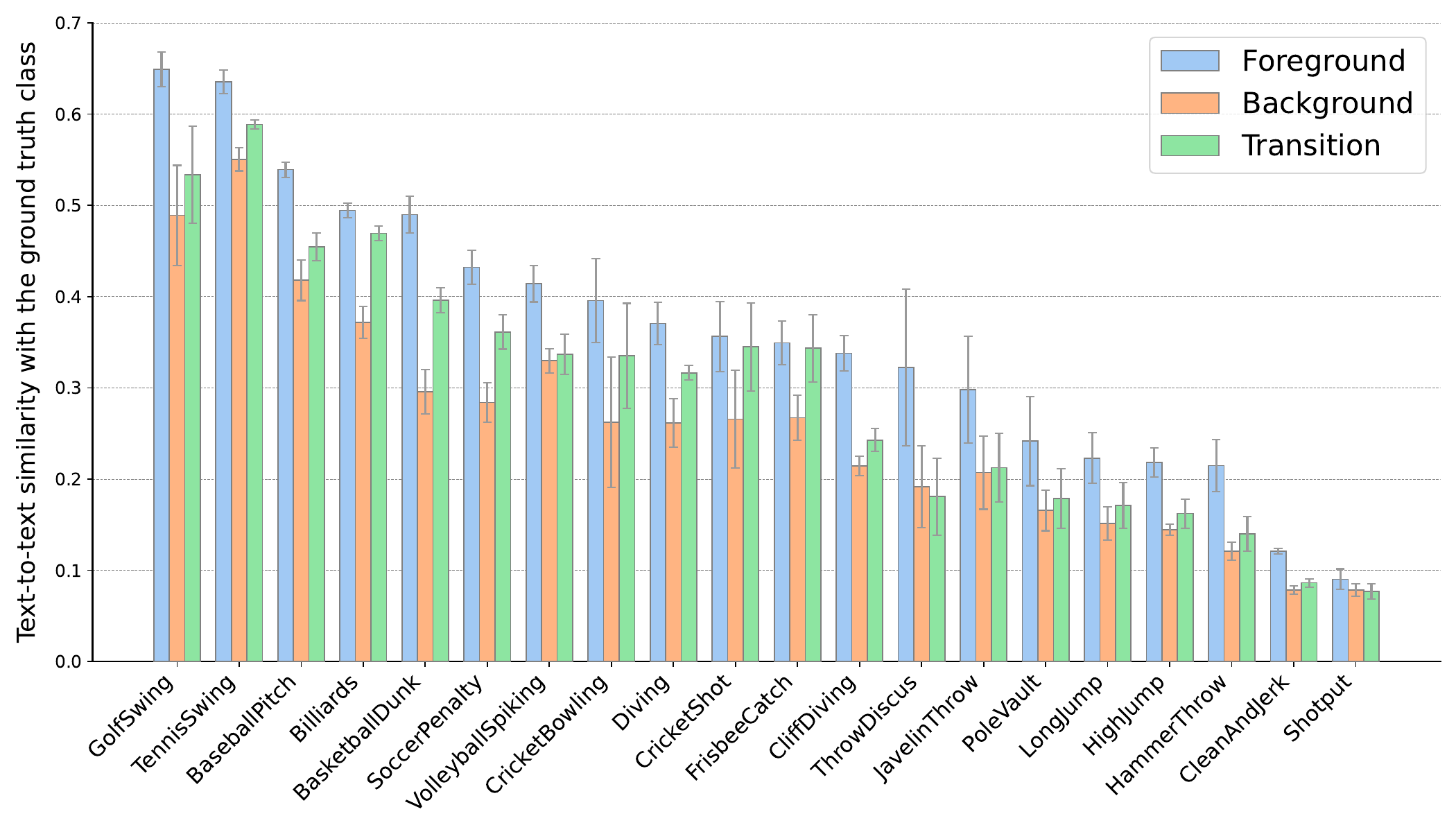}
    \caption{\textbf{Text-to-text similarities on THUMOS14.} Cosine similarities between frame-level captions and ground truth video classes, computed on SentenceBERT embeddings. 
    Numbers are calculated for \inlineColorbox{DrawioBlue}{foreground}, \inlineColorbox{DrawioOrange}{background}, and \inlineColorbox{DrawioGreen}{transition} frames, and averaged across each class. We report the average using different captioning models, together with standard deviations.}
    \label{fig:sim_bert_error_bars}
\end{figure}

\begin{figure*}
    \centering
    \begin{subfigure}[t]{\linewidth}
        \centering
        \includegraphics[width=\linewidth]{supfigures/similarity_comparison_2_v0.pdf}
    \end{subfigure}
    \vspace{0.5cm} 
    \begin{subfigure}[t]{\linewidth}
        \centering
        \includegraphics[width=\linewidth]{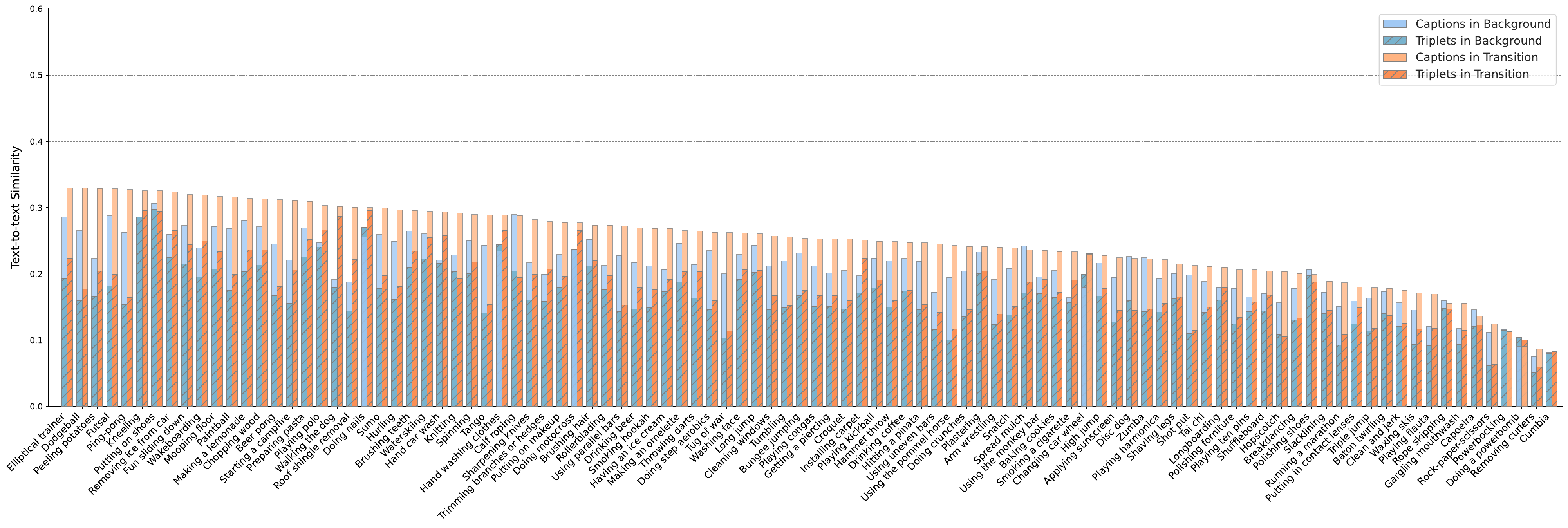}
    \end{subfigure}
    \caption{\textbf{Text-to-text similarities in non-foreground frames.} Cosine similarities between ground truth video classes and (i) frame-level captions grouped by \inlineColorbox{DrawioBlue}{background} and \inlineColorbox{DrawioOrange}{transition} and (ii) scene triplets, again grouped by \inlineColorbox{DrawioBrightBlue}{background} and \inlineColorbox{DrawioBrightOrange}{transition}. Values are averaged per-class.}
    \label{fig:anet_triplets_sup}
\end{figure*}

In \cref{fig:anet_triplets_sup}, as we did for THUMOS14 with Fig.~\refbase{7} in the main paper, we present a comparison between captions and scene triplets.
We compute the similarities between ground truth video classes and both frame-level captions and scene triplets, grouping the results by background and transition frames, and averaging them for each class in ActivityNet-v1.3.
Our findings reveal a clear trend: scene triplets consistently exhibit lower similarity scores in background and transition regions compared to captions,
suggesting that scene triplets possess greater discriminative power in distinguishing relevant content.

\noindent  \textbf{Impact of different captioning models.} 
The results presented in the main paper in Sec.~\refbase{4.3} are not dependent on the specific captioning model used.
By conducting the same analysis with different captioning models, we find that generated captions are generally ineffective at distinguishing between foreground, transition, and background frames. 
Specifically, we utilize BLIP-2~\cite{li2023blip}, CoCa~\cite{yu2022coca}, and KOSMOS-2~\cite{peng2023kosmos} to repeat the analysis shown in Fig.~\refbase{6}, calculating the text-to-text cosine similarities between frame-level captions and ground truth video classes using SentenceBERT embeddings. 
\cref{fig:sim_bert_error_bars} shows the average text-to-text similarities obtained, along with their standard deviations. 
We observe a similar trend, with minimal variability in the similarity values.

%% file: supsec/prompts.tex
\section{Prompts}\label{sec:prompts}
We use the prompts $\mathcal{P}_a$ and $\mathcal{P}_o$ to instruct an LLM to generate detailed action descriptions and the main objects involved in the action, as described in Sec.~\refbase{3}.
The prompt $\mathcal{P}_a$ is defined as follows:
``\textit{You are a helpful assistant with expertise in crafting clear and concise descriptions of human actions. For each request, respond with a list of strings formatted as [" ", " ", ...], without any additional text or explanation. Provide at least one descriptor, and if the action involves multiple objects or steps, generate diverse and relevant descriptions reflecting these variations. Do not include unnecessary adjectives or adverbs. Ensure that each descriptor is atomic, representing a single, distinct action and no longer than 20 words. Only include descriptors for two or more people if the action is typically performed by more than one person. Descriptions should start with 'A person', 'Two people', or 'A group of people' depending on the action.}''.

The prompt $\mathcal{P}_o$ is defined as follows:
``\textit{You are a helpful assistant with expertise in identifying objects associated with specific human actions. For each request, respond with a list of strings formatted as [" ", " ", ...], without any additional text or explanation. Provide at least one object required to perform the action.}''.

%% file: supsec/details.tex
\section{Implementation details}\label{sec:impl}
In this section, we provide additional implementation details that were not included in the main paper. Specifically, we set the margin $m=5$, the factor that weights the contribution of the scene triplets $\alpha=0.5$, the number of centroids for unsupervised clustering $S=20$ for THUMOS14 and $S=5$ for ActivityNet-v1.3. We set $k=2$ for the top-$k$ main action categories. 

%% file: supsec/impact.tex
\section{Potential negative societal impact}\label{sec:impact}
Temporal action localization, akin to other video analysis technologies, is designed to detect human activities in videos. 
This gives rise to potential societal concerns related to privacy, surveillance, and workplace monitoring. 
If misused, TAL could lead to invasive surveillance practices, undermine personal autonomy, or enable unwarranted scrutiny of workers without their consent.
Additionally, misclassification risks in high-stakes scenarios, such as security or law enforcement, could lead to unjust consequences or unfair treatment. 
Responsible deployment and ethical safeguards are essential to mitigate these risks.